\documentclass[ShortAfour]{sage}
\usepackage{graphicx}   % To include figures
\usepackage{amsmath} % assumes amsmath package installed
\usepackage{amssymb}  % assumes amsmath package installed
\usepackage{units}
\usepackage{comment}
\usepackage{color}  % For Highlighting
\usepackage{multirow}
\usepackage{float}
\usepackage[utf8]{inputenc}
\usepackage{gensymb}
\usepackage{array}
\usepackage{epsfig}
\usepackage{graphicx}
\usepackage{float}
\usepackage{subcaption}
\usepackage{enumerate}
\usepackage{mathtools}

\usepackage[round,colon,sort]{natbib}

\usepackage{placeins}

\usepackage{array,booktabs}
\title{\LARGE \bf
Draft: Comparison and Experimental Validation of Predictive Models for Soft, Fiber-Reinforced Actuators}

\author{Audrey Sedal, Alan Wineman, R. Brent Gillespie, C. David Remy% <-this % stops 
\thanks{The authors are with the University of Michigan -- Ann Arbor, Department of Mechanical Engineering.}}

\begin{document}
% ==============================================
% Define additonal paths for including graphics:
\maketitle
%\thispagestyle{empty}
%\pagestyle{empty}

% ==============
\begin{abstract}
Successful soft robot modeling approaches appearing in recent literature have been based on a variety of distinct theories, including traditional robotic theory, continuum mechanics, and machine learning. Though specific modeling techniques have
been developed for and validated against already realized systems, their strengths and weaknesses have not been explicitly compared against each other. In this paper, we show how three distinct model structures ---a lumped-parameter model, a continuum mechanical model, and a neural network--- compare in capturing the gross trends and specific features of the force generation of soft robotic actuators.  In particular, we study models for Fiber Reinforced Elastomeric Enclosures (FREEs), which are a popular choice of soft actuator and that are used in several soft articulated systems, including soft manipulators, exoskeletons, grippers, and locomoting soft robots. We generated benchmark data by testing eight FREE samples that spanned broad design and kinematic spaces and compared the models on their ability to predict the loading-deformation relationships of these samples. This comparison shows the predictive capabilities of each model on individual actuators and each model's generalizability across the design space. While the neural net achieved the highest peak performance, the first principles-based models generalized best across all actuator design parameters tested. The results highlight the essential roles of mathematical structure and experimental parameter determination in building high-performing, generalizable soft actuator models with varying effort invested in system identification.
\end{abstract}
% ==============

%\input{sections/motivation.tex}
%\input{sections/contributions.tex}
%==========================================
\section{Introduction}
\label{sec:motivation}

Soft robots make use of elastic behavior in their constituent materials and therefore often encounter physical phenomena that are neglected in rigid robotic theory. Since soft robots are made from deformable materials and deform themselves, such behaviors include unexpected relationships between expanding and constraining structural elements, as well as deformation- and direction-dependent nonlinear stiffnesses. Accurate soft robot models need new fundamental assumptions that capture multi-dimensional elastic behavior. Further exacerbating the modeling need is the great variety of available soft system designs and control techniques that may require models. Authors publishing under the soft robotics umbrella make use of a broad set of structural schemes at both the actuator level (e.g. cable robots vs. fluid-driven robots) and the system-wide level (e.g. rigid connection between actuators vs. monolithic soft systems), as well as broad choices of materials, models, and functionalities \citep{rus2015design}. Not only is there a modeling need in soft robotics, but there is a problem of model choice.

Successful soft robot modeling approaches appearing in recent literature can be broadly placed into two groups: data-driven strategies and first principle-based strategies. Data-driven strategies have been used on physically realized systems to characterize dynamic behavior, and, in some cases, to develop control policies. Diverse schemes have been proven on a variety of soft systems, including neural networks \citep{giorelli2015neural,giorelli2013feed,gillespie2018learning,thuruthel2018model}, genetic algorithms \citep{giorgio2017hybrid}, and other regression techniques \citep{veale2018accurate,elgeneidy2018bending,bruder2018force}. Within first principles-based models, two subcategories emerge. \citet{della2018dynamic} and others \citep{della2018using,tondu2000modeling,bruder2018force,bruder2017model,satheeshbabu2017designing} show that lumped-parameter models can be used in design and control tasks. Continuum mechanical formulations have recently been demonstrated to be useful in soft system design and control \citep{connolly2017automatic,moseley2016modeling,coevoet2017software,deimel2016novel,goury2018fast,polygerinos2015modeling,uppalapati2018parameter}. First principles-based models also study specific behaviors of soft actuators such as interactions between components \citep{wang2017interaction, tondu2000modeling}.

Many of the models cited above were developed as part of a larger effort to demonstrate capabilities of specific soft systems. Others perform the complementary task of exploring a design space comprised of many possible systems, and determining the optimal design for a given task. Yet, a broad comparison of model structure and features is missing: why, when, and how data-driven models, lumped-parameter models, and continuum mechanical formulations succeed and fail in soft robotics is not well understood.

In this paper, we aim to build this understanding by showing how these three model types compare in capturing the gross behavioral trends and specific features of a popular class of soft actuators. First, we developed three distinct soft actuator models ---a lumped-parameter model, a continuum mechanical model, and a neural network--- that relate the multi-dimensional loading and deformation of this actuator. These models differ in mathematical structure, reaching from simple linear equations to integral expressions to sums of weighted functions. The models also differ in how many parameters they require to be identified from experiment, and in the physical meaning (if any) of the parameters. Next, we generated benchmark data by testing eight soft pneumatically driven actuator samples with varied design parameters, across 22,880 kinematic configurations and pressures. Finally, we compared the models on the gross behavioral features that they were able to capture, and on their performance at predicting kinetics across the design space. 

In particular, we investigated a soft, fiber-reinforced, pneumatic actuator known as the Fiber-Reinforced Elastomeric Enclosure (FREE) \citep{bishop2015design}, or Fiber-Reinforced Soft Actuator (FRSA) \citep{connolly2017automatic}. The FREE consists of a cylindrical, elastomeric tube whose wall is embedded with 1, 2, or 3 families of fibers wound in helices. The helical pitch of these fibers guides the motion that the tube undergoes when internally pressurized. FREEs are used in several existing systems including exoskeletons \citep{koller2016body,polygerinos2015soft,park2014design,singh2018design}, soft manipulators \citep{mcmahan2006field,pritts2004design}, grippers and hands \citep{deimel2016novel}, vibration isolators \citep{scarborough2012fluidic}, bio-inspired slithering systems \citep{branyan2018soft}, and parallel groupings that augment force generation \citep{robertson2017soft}. 

FREEs are a strong candidate for comparative study because they have core mechanical features that are shared in other soft robot architectures. The pressurizeable, nonlinear and hyperelastic wall of the FREE is shared by the PneuNet \citep{shepherd2011multigait} and other soft inflatable robot architectures, while the constraint provided by the fiber is structurally similar to fiber constraints in soft cable-driven robots like the octopus-inspired robot created by \citet{laschi2012soft}. Here, we focus on single fiber-family FREEs because their asymmetric fiber arrangement causes coupling between length change and twisting when the FREE is pressurized \citep{sedal2018continuum,connolly2017automatic}. Together, these key behaviors contrast with the behavior of traditional rigid or series elastic actuators: the FREE design exhibits unexpected relationships between expanding and constraining elements, as well as a nonlinear direction-dependent deformation-loading relationship. FREE models hence demand different governing assumptions than a traditional actuator. Therefore, the FREE serves as a useful investigative platform that captures core problems in soft robotics.

In Section \ref{sec:models}, we present each of the three models and the performance metric on which they are compared. In Section \ref{sec:expt}, we describe the actuator samples and experiments. In Section \ref{sec:results}, we present the experimental results and error analysis of the models. In Section \ref{sec:discussion}, we compare the gross trends and specific features of each model to those of the data, and compare the performance of the models.

%===========================================
\section{Modeling}
\label{sec:models}

\begin{figure}
    \centering
    \includegraphics[width=0.9\linewidth]{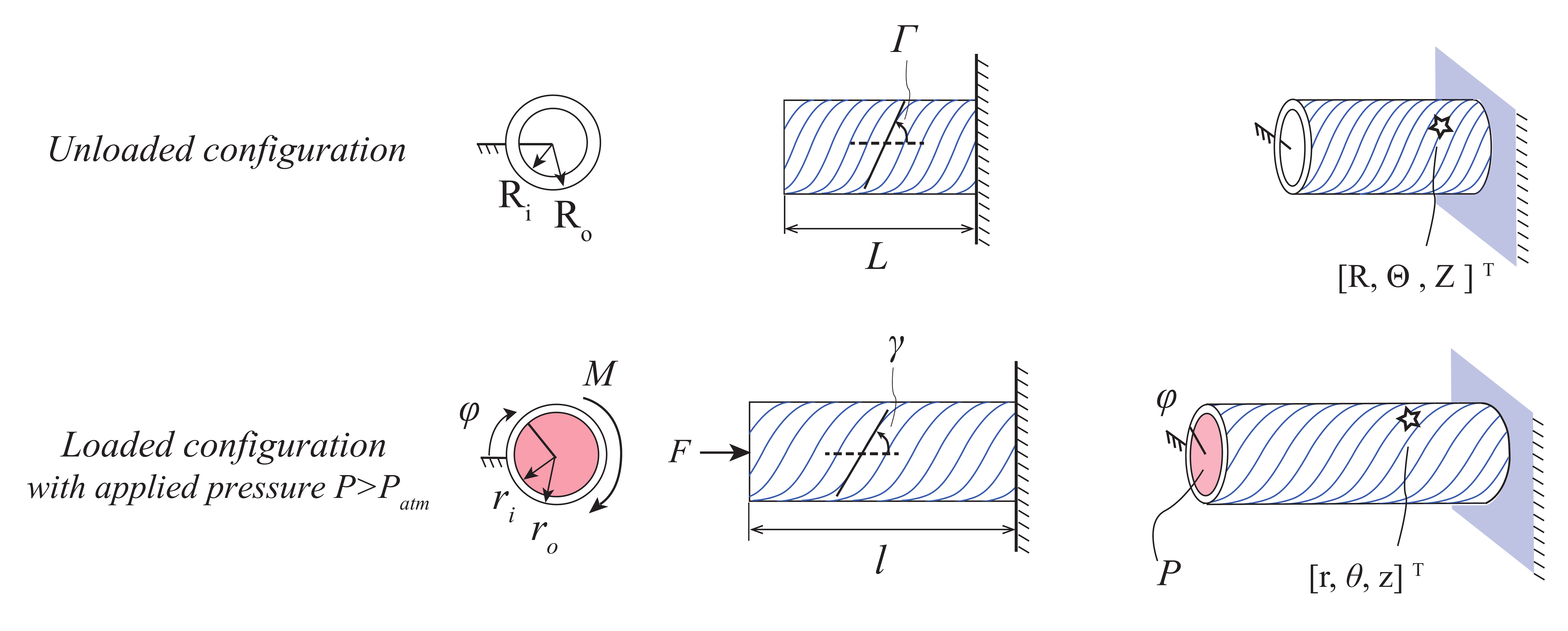}
    \caption{Face, side, and isometric views of a FREE in an unloaded, un-pressurized reference configuration (top) and a loaded, pressurized configuration (bottom).}
    \label{fig:freeCoords}
\end{figure} 

Each model was formulated to relate the FREE's deformation, design parameters, and loading quantities. The unloaded reference configuration of the FREE is a thick-walled, cylindrical tube with length $L$, inner radius $R_i$, and outer radius $R_o$ that is wound with a helical fiber family.  The angle $\Gamma$ between a line tangent to the helix and the axis of the tube defines the geometry of the fiber family (see Figure \ref{fig:freeCoords}). The  set of design parameters $\bar{p} = \begin{bmatrix}\Gamma& L& R_i& R_o\end{bmatrix}$ describes the FREE's unloaded geometry. 

For simplicity, we limited the degrees of freedom (DoFs) under consideration to the FREE's axial elongation and to its end-to-end rotation about the longitudinal axis. This is the intended region of motion of this actuator. The radius is left free to expand or contract. We assumed all other DoFs to be physically constrained. Thus, the current, loaded kinematic state of the FREE is defined by its new length $l$ and its end-to-end twist angle $\varphi$. The vector of generalized coordinates is hence given by $\vec{q} = \begin{bmatrix}l & \varphi \end{bmatrix}^T$. The forces corresponding to the coordinates in $\vec{q}$ are the axial load $F$ and the axial moment $M$, given by the vector of generalized forces $\vec{\tau} = \begin{bmatrix}F & M\end{bmatrix}^T$. An important consequence of this assumption is that the modeled FREE retains its cylindrical shape under any deformation in our choice of kinematic state $\vec{q}$. This assumption is valid for cylindrical FREEs which have not undergone a buckling instability. It is also a good approximation for a lightly bent actuator whose bending radius greatly exceeds its cross-sectional radius. Then, for a FREE with a given set of design parameters $\bar{p}$, the goal of our modelling effort is to characterize the relationship between the kinematic state $\vec{q}$ and the generalized force output $\vec{\tau}$, and to express this relationship as a function of the applied fluid pressure $P$.

Without loss of generality, we chose to formulate this relationship in terms of a force prediction problem and sought to develop and compare different types of models that establish the following functional relationship:
\begin{equation}
            \vec{\tau} = \boldsymbol{f}(\vec{q},P,\bar{p}).
        \label{eq:mainProbStmt}
\end{equation}
The inverse problem of finding an actuator's kinematic state $\vec{q}$ given its generalized forces $\vec{\tau}$ and design parameters $\bar{p}$ may be solved numerically, in the case of first principles models, or by re-training, in the case of data-driven models. 
The three models presented in the following sections differ significantly in their mathematical structure and in the number of free parameters that must be identified through experimental data. The physical meaning (if any) of these parameters is also discussed.

\subsection{Linear Lumped-Parameter Model}
\label{sec:lpm}
The primary assumption of our first model is that fibers of the FREE are inextensible and create a perfect kinematic constraint forming a helix that encloses the elastomeric tube.
Considering only the unbuckled case in which the FREE retains its cylindrical shape, we use this assumption to derive an analytic expression for the outer radius $r_o$ (Fig. \ref{fig:freeCoords}) as a function of the state $\vec{q}$ and the design parameters $\bar{p}$ \citep{bruder2018force}:
\begin{equation}
    r_o \left(\vec{q}, \bar{p}\right) = \frac{ \sqrt{B^2 - l^2}}{\left|\Phi + \varphi\right|}.
    \label{eq:linearRadius}
\end{equation}
Here, the fiber length $B$ is given by $B = L \frac{1}{\cos{\Gamma}}$ and the initial wrapping angle $\Phi$ by $\Phi = \frac{L}{R_o} \tan{\Gamma}$.
 
Neglecting the wall thickness of the FREE (i.e., assuming that the inner radius $r_i \approx r_o$), we can employ this expression to compute the volume $V$ of the fluid inside the FREE:
\begin{equation}
    V\left(\vec{q}\right) = \pi l r_i^2 = \pi \frac{l B^2 - l^3}{(\Phi + \varphi)^2}. 
    \label{eq:linearVolume}
\end{equation}
Taking the partial derivative of this expression with respect to the kinematic state $\vec{q}$ yields the fluid Jacobian $\textbf{J}_V$:
\begin{equation}
    \textbf{J}_V =  \frac{\partial V}{\partial \vec{q}}
                 =  \begin{bmatrix} 
    		            \pi \frac{B^2 - 3l^2}{(\Phi + \varphi)^2} 
    		            & 
    		            - 2 \pi \frac{l B^2 - l^3}{(\Phi + \varphi)^3}
    		        \end{bmatrix}.
    \label{eq:fluidJ}
\end{equation}
The transpose of this Jacobian allows us to compute the generalized forces $\vec{\tau}_{fluid}$ that are created by the fluid pressure $P$ \citep{bruder2018force}:
    
\begin{equation}
    \vec{\tau}_{fluid} = \textbf{J}_V^T P.
    \label{eq:linearFluidForces}        
\end{equation}
    
These fluid forces are summed with the elastic forces that result from the deformation of the FREE wall in axial compression, axial twist, and radial compression:
\begin{equation}
    \vec{\tau} =  \vec{\tau}_{fluid} + \vec{\tau}_{wall}.
    \label{eq:linearSumForces}        
\end{equation}
The second major assumption of this model is that the FREE wall provides a linear elastic response to deformations in $\vec{q}$. Using $\Delta \vec{q} = \begin{bmatrix} l-L& \varphi\end{bmatrix}$ for the deformation along the generalized coordinates, we can compute these elastic forces as:

\begin{equation}
    \vec{\tau}_{wall} = -\textbf{K}  \Delta \vec{q} = - \begin{bmatrix} k_{a}& k_c \\ k_c& k_{b}\end{bmatrix} \Delta \vec{q}.
\end{equation}

The stiffness matrix $\textbf{K}$ is a positive-definite, symmetric matrix that is found by fitting model parameters $k_a$, $k_{b}$, and $k_c$ to experimental data.
The diagonal elements in this matrix, $k_a$ and $k_{b}$, approximately correspond to the lumped stiffness of the wall in axial compression and twist, respectively. 
Because of the fibers in a FREE, it is necessary to also include off-diagonal elements $k_c$ in $\textbf{K}$.
As expressed in Eq.~\eqref{eq:linearRadius}, motion in $l$ and $\varphi$ induces a change in the radius of the FREE and hence a radial compression of the wall.  
The elastic response to this compression is transferred by the fibers back into the generalized torques $\vec{\tau}$.
This effect causes an elastic coupling between the twist and axial directions, resulting in the off-diagonal terms in $\textbf{K}$.

We can express the complete lumped parameter model for a FREE by combining these expressions:
\begin{equation}
\vec{\tau} = \textbf{J}_V^T  P - \textbf{K} \Delta \vec{q},
    \label{eq:LumpedFormMain}
\end{equation}
which is the special form of Eq. \ref{eq:mainProbStmt} for the linear lumped-parameter model. 

For this particular model, it is worth noting that the radius $r_o$, the volume $V$, and the fluid Jacobian may have singular terms when $\varphi = -\Phi = -\frac{L}{R_o} \tan \Gamma$. 
This deformation corresponds to a configuration in which the actuator has been rotated such that the fibers are parallel to the tube's central axis. In this case, the radius and internal volume become ill-defined and the fibers can store any amount of tension or compression in axial directions. In this formulation, the model parameters that must be fit experimentally are the three parameters of lumped stiffness $k_a$, $k_{b}$, and $k_c$.

\subsection{Nonlinear Continuum Model}
\label{sec:cont_model}

In our second model, we predict the generalized forces $\vec{\tau}$ in Equation~\eqref{eq:mainProbStmt} with a continuum-based, non-linear relationship. Though we continue to assume that the FREE is a cylindrical tube, its wall thickness is no longer neglected and the fiber is considered to be extensible. Extensible fibers cannot kinematically define the tube's geometry; the radius and internal volume of the FREE in this model depend not only on the generalized coordinates $\vec{q}$, but also on the internal pressure $P$. Finally, we replace the assumption of a linear elastic response and replace it with a nonlinear response arising from consideration of the wall's deformation and occupied volume.

To establish %maintain 
a well-defined problem, we %instead 
assume that:
\begin{enumerate}
    \item The actuator deforms from a thick-walled tube to a thick-walled tube of different dimensions,
    \item the volume occupied by the material of the FREE wall stays constant (i.e. the FREE wall is incompressible, as evidenced for rubbers according to \citet{gent2012engineering}), and
    \item the fiber is perfectly embedded into the elastomer and evenly distributed through it, such that the interaction phenomena are homogenized throughout the FREE wall and the fiber remains locally tangent to the elastomeric tube.
\end{enumerate}

These assumptions enable a modified implementation of a continuum mechanical framework proposed by \cite{holzapfel2000new}. This specific model implementation has been presented before \citep{sedal2018continuum} and is summarized here for completeness.

Under the assumptions listed above, the FREE transforms from a tube with initial dimensions $\{L, R_i,R_o\}$ for length, interior radius, and exterior radius, and no end-to-end twist ($\varphi = 0$) to a tube with new dimensions $\{l,r_i, r_o\}$ and end-to-end twist $\varphi$ (Figure \ref{fig:freeCoords}). 
%Then, the point in the reference configuration $\vec{X}= \left[R \; \Theta \; Z\right]^T$ undergoes a change of coordinates to $\vec{x} = \left[r\; \theta \; z\right]^T$ 
To define this deformation, we track an arbitrarily chosen elemental volume in the FREE wall. The elemental volume has location coordinates $\vec{X} = \left[R \; \Theta \; Z\right]^T$ in the FREE's load-free configuration, and coordinates $\vec{x}  = \left[r\; \theta \; z\right]^T$ in the current, loaded configuration such that $\vec{x} = \textbf{g}(\vec{X})$. The following functions define the new coordinates $\vec{x}= \textbf{g}(\vec{X})$ of any point $\vec{X}$ in the unloaded configuration:

\begin{align}
        r = \sqrt{\frac{R^2 - R_i^2}{ \lambda_z}+r_i^2} , \\
        \theta = \Theta + Z\frac{\varphi}{L} , \\
        z = \lambda_z Z, \\
        %\lambda_z \triangleq \frac{l}{L}.
        \textrm{where } \lambda_z \coloneqq \frac{l}{L}.
        \label{eq:nlCoordChange}
\end{align}

The deformation gradient \textbf{F}, which describes the deformation of the particles in the FREE, is defined as:
\begin{equation}
\textbf{F} = \left[\begin{array}{ccc} 
 \frac{\partial r}{\partial R} 
 & \frac{\partial r}{ R \partial \Theta} 
 & \frac{\partial r}{\partial Z} \\
  \frac{r \partial \theta}{\partial R} 
 & \frac{r \partial \theta}{R \partial \Theta} 
 & \frac{r \partial \theta}{\partial Z} \\
 \frac{\partial z}{\partial R} 
 & \frac{\partial z}{R \partial \Theta}
 & \frac{\partial z}{\partial Z} 
 \end{array} \right] = 
 \left[\begin{array}{ccc} 
 \frac{R}{r\lambda_z} & 0 & 0 \\
 0 & \frac{r}{R} & r \frac{\Phi}{L} \\
 0 & 0 & \lambda_z\
 \end{array} \right]. \\ 
\label{eq:defGrad}
\end{equation}

In Holzapfel's framework, composite materials are modeled by considering the ``strain energy," i.e. Helmholtz free energy, stored in the deformed material.

The previously listed assumptions allow us to write the strain energy $\Psi_{total}$ of the FREE wall as a superposition of the free energy from the elastomer and the free energy from the fiber:
\begin{equation}
    \Psi_{total} = \Psi_{isotropic} + \Psi_{anisotropic}.
    \label{eq:strainEnergySuperposition}
\end{equation}
Here, each free energy depends on the deformation gradient \textbf{F} so as to represent particular material behavior: the fiber is a ``standard" fiber that is anisotropic in space \citep{holzapfel2000new,gent2012engineering} and the elastomer is an isotropic neo-Hookean solid \citep{ogden1997non}.
The neo-Hookean solid model introduces a material parameter $C_1$ and an invariant quantity $I_1$ % is determined from \textbf{F} 
associated with the elastomer's isotropy \citep{spencer1971part}, such that:
\begin{equation}
    \Psi_{isotropic} = \frac{C_1}{2} \times (I_1(\textbf{F}^T\textbf{F}) - 3),
    \label{eq:neoHook}
\end{equation}
The standard fiber model gives the energy associated with the fiber anisotropy by introducing a material parameter $C_2$ and an invariant quantity $I_4$ associated with the fiber stretch \citep{spencer1971part}:

\begin{equation}
    \Psi_{anisotropic} = \frac{C_2}{2} \times (I_4(\textbf{F}^T\textbf{F},\Gamma) - 1)^2.
    \label{eq:standardFiber}
\end{equation}

\begin{figure}
    \centering
    \includegraphics[width=0.25\linewidth]{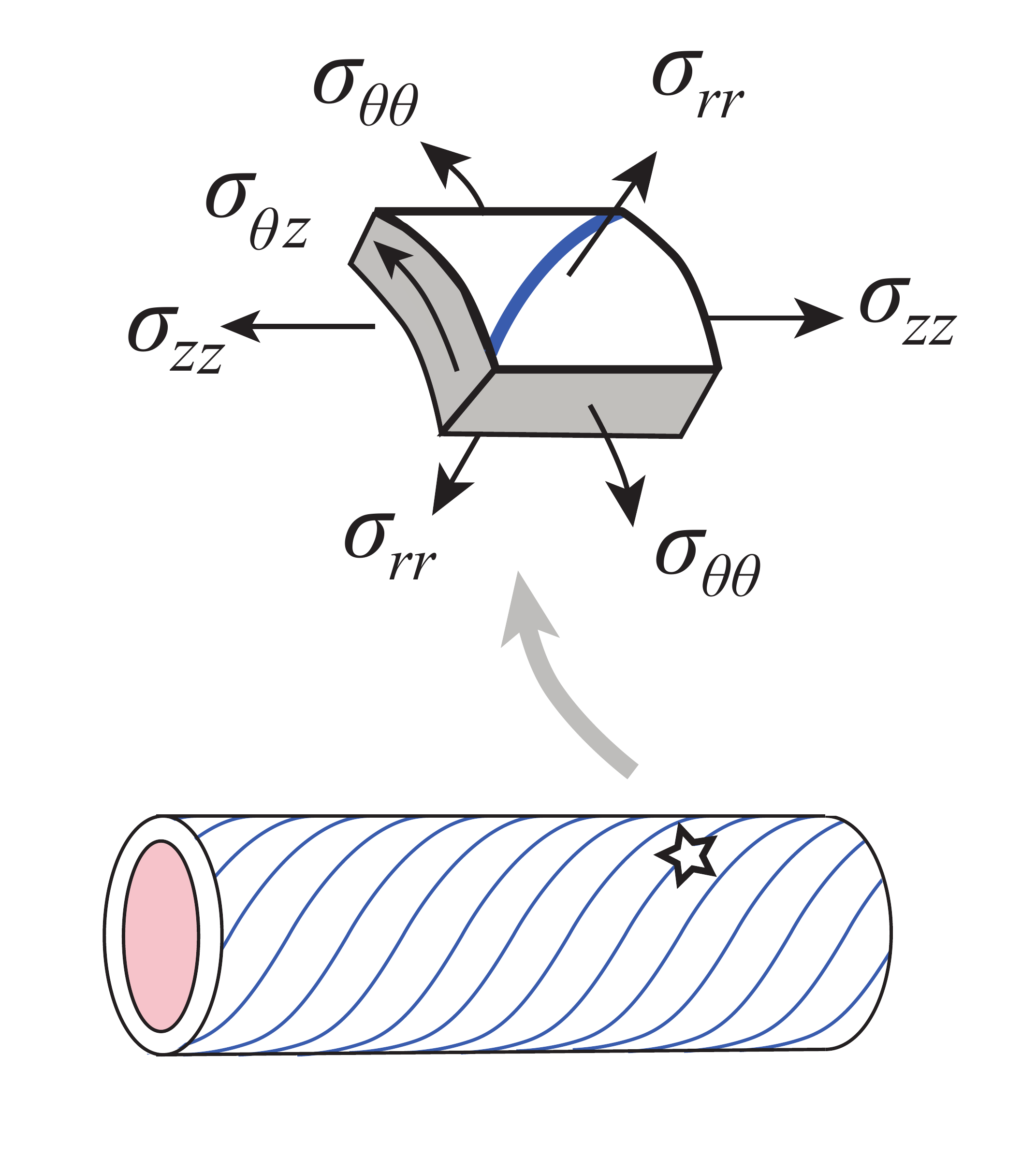}
    \caption{Arbitrary elemental volume of the FREE wall (marked by a star) with relevant stresses shown.}
    \label{fig:stresses}
\end{figure}

The stresses (i.e. internal forces per area) inside the FREE wall are found by taking the derivative of the total Helmholtz free energy $\Psi_{total}$ with respect to the deformation gradient \textbf{F} and including a Lagrange multiplier $b$ that accounts for the FREE wall incompressiblity. The expression for three-dimensional stress stored in the deformed FREE wall is then:

\begin{equation}
    \boldsymbol{\sigma} = \left[\begin{array}{ccc} 
    \sigma_{rr} & \sigma_{r\theta} &  \sigma_{rz}\\
    \sigma_{\theta r} & \sigma_{\theta\theta} & \sigma_{\theta z} \\
    \sigma_{zr} & \sigma_{z \theta} & \sigma_{zz}
    \end{array} \right] = \frac{\partial \Psi_{total}}{\partial \textbf{F}} \textbf{F}^T - b \textbf{I}.
    \label{eq:nlStress}
\end{equation}

Using the continuum approach enables a system of partial differential equations for the stresses which express the equilibrium of the material under $\vec{\tau}$. In cylindrical coordinates and under the symmetry of the FREE, it is possible to manipulate these equations to develop an expression for $b$. See the appendix of \citet{sedal2018continuum} for a detailed derivation. Since the FREE's radii $r_i$ and $r_o$ are unconstrained, we use the same equations of equilibrium in \citet{sedal2018continuum}, subject to the boundary condition of the FREE's internal pressure $P$, to solve for $r_i$:
\begin{equation}
      -P = \int_{r_i}^{r_o} \frac{1}{r} ( \sigma_{rr} - \sigma_{\theta \theta} )dr 
    \label{eq:intRad}
\end{equation}
where $r_o$ is found using the material's incompressibility:
\begin{equation}
     r_o = \sqrt{r_i^2 + \frac{R_o^2-R_i^2}{\lambda_z}}.
     \label{eq:volcons} \\
\end{equation}

Having established $r_i$ and $r_o$, we compute the axial force and moment on the FREE by integrating the stresses over the radius of the FREE:
\begin{equation}
F = -2 \pi \int_{r_i}^{r_o} \sigma_{zz} r dr + \pi r_i^2 P
    \label{eq:force}
\end{equation}
\begin{equation}
 M =  -2 \pi \int_{r_i}^{r_o} \sigma_{\theta z} r^2 dr.\\
\label{eq:moment}
\end{equation}
Here, the stresses are expressed in terms of $\bar{p}, \vec{q}$, and $P$ through the use of Eqs. \eqref{eq:nlStress}, \eqref{eq:strainEnergySuperposition}, and \eqref{eq:defGrad}. This establishes a nonlinear expression:
\begin{equation}
    \vec{\tau} = \boldsymbol{f}_{c}(\vec{q},P,\bar{p}).
\end{equation}
In this formulation, the model parameters that must be fit experimentally are the two material parameters $C_1$ and $C_2$ which represent the stiffness of the elastomer and fiber. 

\subsection{Neural Network Model}
\label{sec:nnmodel}
\label{sec:nn_model}

In our third model, we predict the generalized forces in Eq.~\eqref{eq:mainProbStmt} with a data-driven approach using a neural network.
Our network implementation is based on the previous success of neural networks in modeling the kinetics \citep{giorelli2013feed} and statics \citep{giorelli2015neural} of structurally similar robots.
In particular, we implemented the neural network presented by \citet{giorelli2015neural} using the inputs and outputs of the FREE statics problem as defined in Eq.~\eqref{eq:mainProbStmt}. We refer to this publication for a more detailed description of the neural network, but summarize our implementation below for completeness.

\begin{figure}
    \centering
    \includegraphics[width=3in]{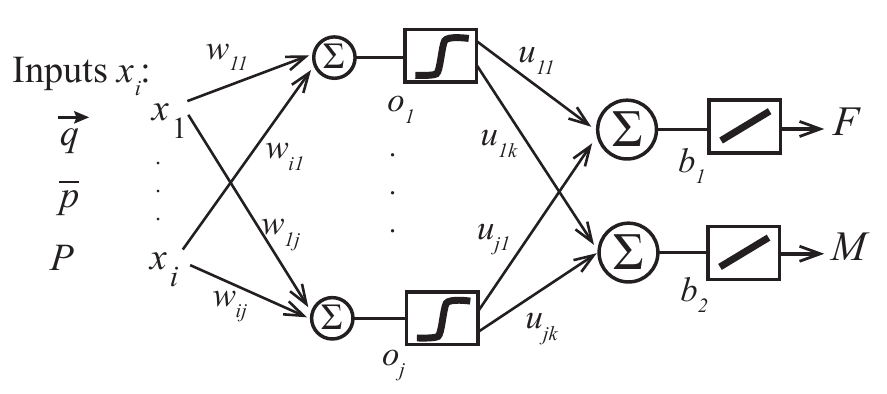}
    \caption{Schematic of the neural network, with inputs $vec{q},\bar{p}$ and $P$ ($x_i$ for $i \in \{1,...,7\}$), hidden layer neurons $o_j$ with $j \in \{1,...,6\}$ and outputs $F$ and $M$.}
   \label{fig:nnschematic}
\end{figure}

The set of inputs to our neural network consisted of the kinematic state $\vec{q}$, the internal pressure $P$, and the design parameters $\bar{p}$ of each sample. The outputs of the neural network were the components $F$ and $M$ of the generalized forces $\vec{\tau}$.
This neural network was implemented as a shallow network with a single fully connected hidden layer containing 6 neurons using a hyperbolic tangent activation function. A schematic of the neural net with inputs $x_i$, weights and biases $w$, $u$, $o$ and $b$, and outputs $F,M$ is shown in Fig. \ref{fig:nnschematic}.

The weights $w_{i,j}$ and $u_{j,k}$, and biases $o_j$ and $b_{1,2}$ were fit experimentally through training with a back propagation algorithm.
With our chosen network topology, 7 inputs, 2 outputs, and 6 neurons on a fully-connected hidden layer, the network had a total of 42 weights $w_{i,j}$, 12 weights $u_{j,k}$, 6 biases $o_j$ and 2 biases $b_{1,2}$. This gives a total of 62 parameters that must be determined experimentally. However, unlike in the other models, these parameters have no physical interpretation.

%===============================================================
\section{Hardware Experiments} %maybe change name
\label{sec:expt}

To determine the predictive ability of each model, we performed a suite of experiments on a set of eight FREE samples spanning the design space %in $\Gamma$ 
under various loading conditions and imposed kinematic states. 

\subsection{Samples}

All sample FREEs were made from cotton thread (``Aunt Lydia's'', Size 10) adhered to rubber tubing. Natural rubber tubing with specified \unit[9.5 $\pm$ 0.25]{mm} inner diameter, \unit[1.6 $\pm 2\times 10^-1$]{mm} thickness (Kent Elastomer) was coated with a thin layer of rubber cement (Elmer's), resulting in sample thickness between 1.3 and \unit[2]{mm}. During winding, the fiber angle was prescribed by a 3D-printed template inserted into the rubber tube. Fiber spacing was \unit[1.67 $\pm$ 0.25]{mm}. 
After winding, a thin layer of liquid latex (TAP) was applied by hand to further secure the fibers on the tube. Samples were cut between 8 and \unit[12]{cm} in length.%, and had between  of thickness.

We created eight samples with fiber orientations $\Gamma$ spanning the design space $\Gamma \in (0\degree, 90\degree)$ in increments of roughly $10 \degree$. Fiber orientation of the finished samples was measured through a photograph in three locations and averaged. Standard deviation of the measured fiber orientation did not exceed $1 \degree$. Though $\Gamma$ is the main design variable, our samples also differed somewhat in length and wall thickness. The dimensions of length and thickness were measured with a micrometer three times on each FREE sample and averaged. Standard deviation of sample length and thickness measurements did not exceed \unit[0.10]{mm} and \unit[0.13]{mm} respectively. Table \ref{table:fiber_angles} shows the fiber orientation and initial dimensions of each sample.% Each uncertainty shown is the maximum observed standard deviation between three measurements on any sample, except for inner diameter uncertainty, which is the manufacturer tolerance.

%Though templates guided the fiber winding process, there was still variation between the intended fiber angle and the average fiber angle of the sample. For this reason, 

\begin{table}[]
    \centering
\begin{tabular}{c | c  c  c  c  c  c  c  c}
    Sample & 1 & 2 & 3 & 4 & 5 & 6 & 7 & 8\\
    \hline
    $\Gamma$ ($\degree$) & 15 & 25 & 36 & 40 & 50 & 62 & 73 & 76\\
    $L$ (mm) & 90.48 & 120.52 & 98.42 & 90.48 & 120.40 & 99.00 & 128.9 & 103.22\\
    $R_i$ (mm) & 4.77 & 4.77 & 4.77 & 4.77 & 4.77 & 4.77 & 4.77 & 4.77\\
    $R_o$ (mm)  & 6.13 & 6.62 & 6.74 & 6.13 & 6.41 & 6.36 & 6.40 & 6.18 \\
\end{tabular}
\caption{Design parameters $\bar{p}$: fiber orientation $\Gamma$ ($\degree$) and initial dimensions of length $L$, inner diameter (ID), and wall thickness $t$ of each FREE sample.}
\label{table:fiber_angles}
\end{table}

\begin{comment}
\begin{table}[]
    \centering
\begin{tabular}{c | c  c  c  c  c  c  c  c}
    Sample & 1 & 2 & 3 & 4 & 5 & 6 & 7 & 8\\
    \hline
    $\Gamma$ ($\degree$) & 15 & 25 & 36 & 40 & 50 & 62 & 73 & 76\\
    $L$ (mm) & 90.48 & 120.52 & 98.42 & 90.48 & 120.40 & 99.00 & 128.9 & 103.22\\
    ID (mm) & 9.53 & 9.53 & 9.53 & 9.53 & 9.53 & 9.53 & 9.53 & 9.53\\
    $t$ (mm)  & 1.36 & 1.85 & 1.97 & 1.36 & 1.64 & 1.59 & 1.63 & 1.41 \\
\end{tabular}
\caption{Design parameters $\bar{p}$: fiber orientation $\Gamma$ ($\degree$) and initial dimensions of length $L$, inner diameter (ID), and wall thickness $t$ of each FREE sample.}
\label{table:fiber_angles}
\end{table}
\end{comment}

After fabrication, the samples were fit using %2 or 3 
parallel zip ties to the barbed side of \unit[9.5]{mm} (\unit[$3/8$]{in}) single barb to ``\unit[$1/8$]{in} NPT'' style pneumatic fittings.

\subsection{Testing Platform}

\begin{figure}
        \centering
        \includegraphics[width=3in]{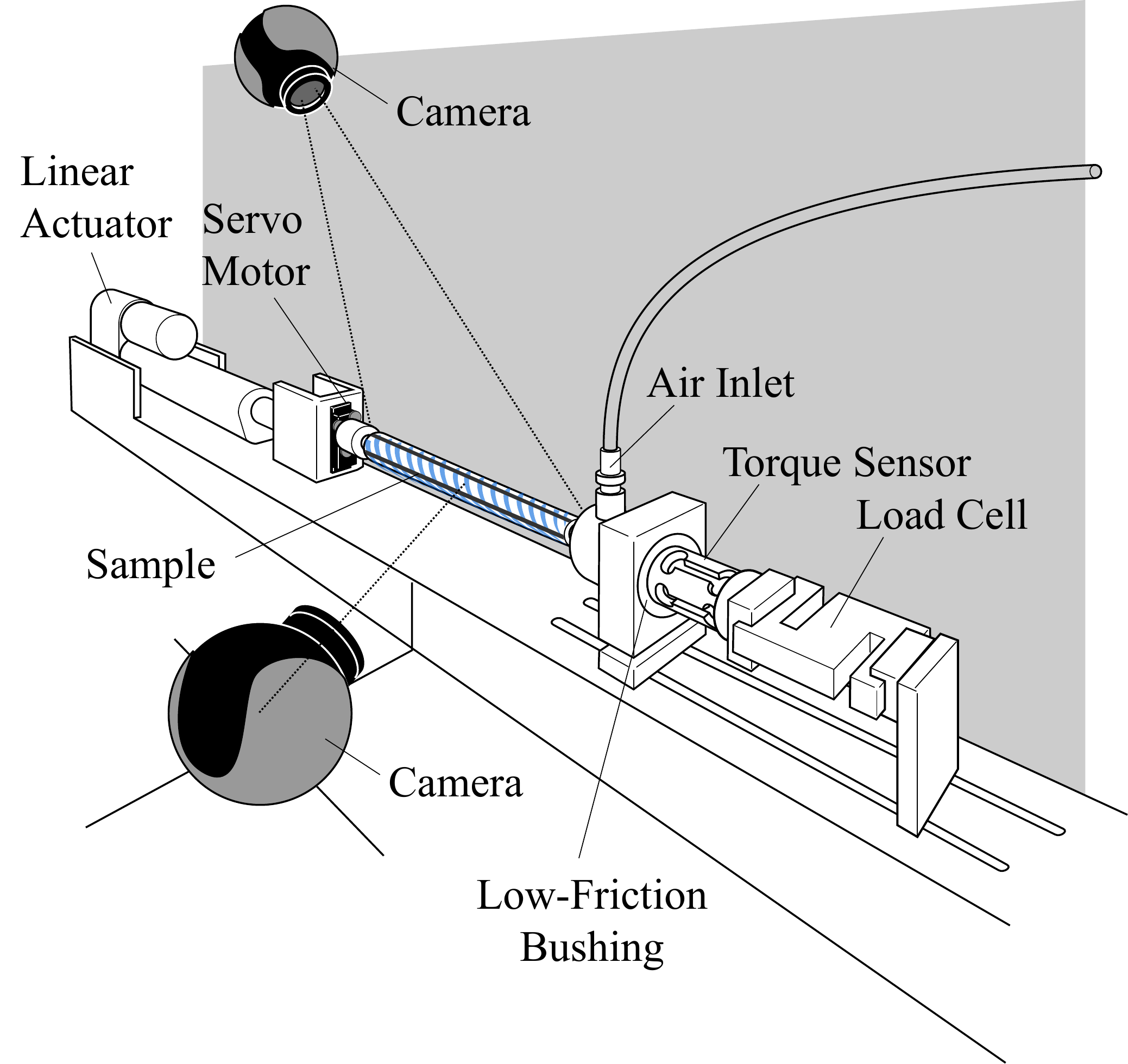}
        \caption{Test bed. The sample FREE, shown with blue fibers, is mounted in the center. On the left, the linear actuator and servo motor fix the length and twist. On the right, the torque sensor and load cell measure the loads while the FREE is inflated via the air inlet through a flexible, lightweight tube. Between the air inlet and the sensors is a cylindrical teflon bushing. Top and side camera take low resolution video and high resolution photos.}
        \label{fig:testrig2}
\end{figure}

Each sample was fitted into a custom-built testing platform (Figure \ref{fig:testrig2}) designed to elongate, twist and pressurize the samples while measuring loading at the tip and photographing the sample's outer wall.

\subsection{Testing Protocol}

Prior to testing, test bed error was characterized. The linear actuator and rotational servo were tested to ensure position control capabilities within \unit[0.146]{mm} and 0.35$\degree$ respectively. Cross-talk between the load cells was measured using known forces and torques between 0 and \unit[10]{N} and 0 and \unit[100]{Nmm}. Error due to cross-talk on the force sensor was \unit[3.43E-3]{ N/Nmm}, and the same error on the torque sensor was \unit[6E-1]{Nmm/N}. Parasitic friction on the platform was measured by cycling known tensile loads between 0 and \unit[7]{N}, and was found not to exceed sensor resolution.

Each FREE sample was mounted into the test bed using custom NPT thread attachments and teflon tape.
Aligned markings were placed on the FREE sample and mounting points to observe potential slippage.
Then, each sample was tested for all possible combinations of the following kinematic states $\vec{q}$ and internal pressures $P$:

\begin{equation}
   % \begin{align}
        \Delta l \textrm{(mm)} = \{-5, -4, ...,-1, 0, 1, ..., 4, 5\} \\
        \label{eq:testParamSets00}
\end{equation}
\begin{equation}
        \Delta \varphi (\degree) = \{-120, -110, ..., -20, -10, -1, 1, 10, 20, ..., 110, 120 \} \\
        \label{eq:testParamSets10}
\end{equation}
\begin{equation}
        P_{in} (\textrm{V}\times 10^{-1}) = \{0, 1, 2, 3, 4, 5, 6, 7\}
    \label{eq:testParamSets20}
   % \end{align}
\end{equation}

Iterating through these configurations, we first commanded the rotational servo and linear actuator to create a desired end-to-end rotation and axial stretch/compression. As shown by Eqns.~\eqref{eq:testParamSets00}-\eqref{eq:testParamSets20}, each sample was tested in 286 distinct kinematic states $\vec{q}$.
Then, gauge pressure was set by a voltage signal that corresponded to pressures between 0 and \unit[72.5]{kPa}. Each sample was inflated to a control signal for \unit[72.5]{kPa} in eight steps, and deflated back to atmospheric pressure in two additional steps. This resulted in 2,860 configurations tested per sample. After the command pressure was reached, each of these configurations was held for 20 seconds while data (force, torque, and pressure) were collected at \unit[1]{Hz}.
This data was synchronized and averaged to yield a single measurement triplet of $\vec{\tau}$, $\vec{q}$, and $P$. 
Throughout each test, the top camera took time-stamped video at 15 fps to capture any unforeseen events. After the 20 seconds, the side camera photographed the FREE sample wall and then the command for the next configuration was sent. 
Each photo was later manually classified for whether the FREE sample had buckled in the given configuration or not.
All buckled samples were excluded from further analysis.
Over all eight samples, loading data and images were collected for 22,880 configurations. 
After testing was completed, samples were inspected and it was verified that none had been damaged throughout the test.

\subsection{Model Comparison Procedure}

\label{sec:modelEvals}

Parameter fitting and model evaluation were done on the separated data sets. We randomly partitioned the data obtained from the un-buckled configurations of each sample into a training set ($80\%$ of the data) and test set ($20\%$ of the data). Thus, there were eight distinct, randomly chosen and randomly ordered training sets, and eight distinct and similarly random test sets. We then created two additional training-test pairs by aggregating the training and test data for the even-numbered samples (Samples 2, 4, 6, and 8) and for the data from all the samples.

\subsubsection{Error Metric} 
As an aggregated error metric across a set of $n$ data, we expressed the model error $E$ as the root mean square error (RMSE) of force and moment errors normalized by their maximal measured values $F_{meas}^{max}$ and $M_{meas}^{max}$, respectively.  This error metric was used to fit model parameters to training data and to determine model performance against test data:

\begin{align}
    E = \sqrt{
        \frac{1}{n} \sum_{i=1}^{n} \left(\frac{{F}_{meas}^i-{F}_{model}^i}{F_{meas}^{max}}\right)^2 
        +
        \left(\frac{{M}_{meas}^i-{M}_{model}^i}{M_{meas}^{max}}\right)^2
    }
    \label{eq:errorObj}
\end{align}

\subsubsection{Model Parameter Identification} 

 Model parameters were fit to training data by minimizing the error metric $E$ (Eq.~\eqref{eq:errorObj}). The fitting operation was undertaken for each of the three models individually on all ten training data sets. Each individually trained model, with its corresponding parameter values, was then used to predict the relationship between loading, pressure, and kinematic state for each test data set. Predicted values were then compared with the measured values to determine model performance using $E$. For the continuum model, we additionally calculated $E$ on all test sets using model parameters gathered from tests on isolated materials. 

The three model parameters $k_a$, $k_b$, and $k_c$ of the linear lumped model were fit through a constrained minimization of $E$ over a given training data set.
The required positive-definiteness of the stiffness matrix \textbf{K} was implemented as a positivity constraint on the eigenvalues of \textbf{K}.
The constrained optimization problem was solved with a gradient based interior-point algorithm (\textit{fmincon}) in Matlab.
The algorithm was initialized at order of magnitude estimates of $k_a = k_b =k_c = 1$.

In a similar fashion, we found the two parameters of the nonlinear continuum model, using the positivity constraint $C_{1,2}>0$.
The initial estimates for $C_1$ and $C_2$ were $10^5$ and $10^6$ Pa respectively. These are order of magnitude estimates based on previous physical measurements of rubber and cotton fibers \citep{gent2012engineering,sedal2018continuum}.
The constrained optimization problem was again solved using \textit{fmincon}.

Since the constants $C_1$ and $C_2$ represent actual physical properties of the material used for the wall and the fiber, they were also identified via experiments on small material samples.  In addition to fitting $C_{1,2}$ from composite sample measurements as described above, we also evaluated the continuum model's performance using individual constituent values of $C_{1,2}$ of the elastomer and fiber respectively from \citet{sedal2018continuum}.

The 62 parameters of the neural network were fit with back propagation on normalized input values. We performed training using an unconstrained gradient based method (Levenberg-Marquardt algorithm), implemented with Matlab's \textit{train} function. Unlike the other models, the neural network requires the data to be paritioned into 3 sets. Thus, we used $64 \%$ for training, $16 \%$ for validation, and $20 \%$ for testing. Up to 1000 epochs were permitted. The training of the neural network was based on $E^2$ as objective function, which has the same global minimum as $E$.

%=====================================================

\section{Results}
\label{sec:results}
\subsection{Buckling}
\label{sec:bucklingResults}

\begin{figure}
\centering
\label{fig:photoBuckling}
    \includegraphics[width=6in]{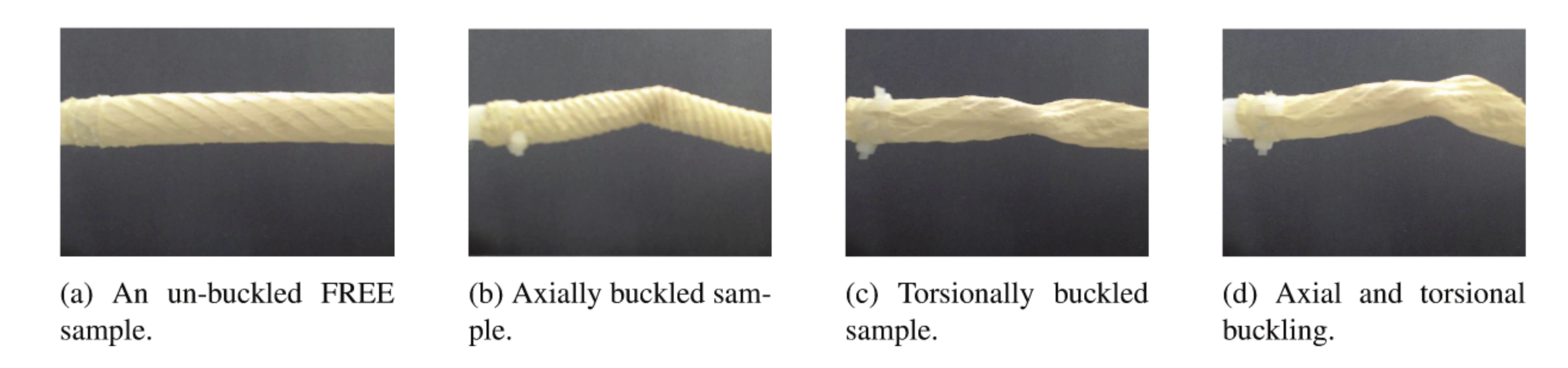}
\caption{Images from experiment of sample configurations: (a) un-buckled, (b) axially buckled, (c) torsionally bucked, and (d) both axially and torsionally buckled.}
\end{figure}

Out of the 22,880 tested configurations, 12,575 conditions (55.0\%) were unbuckled (Figure \ref{fig:buckledIm3}), while  10,305 (45.0\%) showed signs of either axial buckling (Figure \ref{fig:buckledIm1}), torsional buckling (Figure \ref{fig:buckledIm2}), or both (Figure \ref{fig:buckledIm4}).
%Axial buckling was characterized by ...
%Torsional buckling was characterized by  ...
Since the geometric assumptions of Section \ref{sec:models} are broken in buckled FREEs, they were not included in subsequent analysis. The remaining configurations are shown in Figure 6. It is not too surprising that the FREEs buckled under several tested conditions: the broad, standardized set of kinematic states and pressures imposed on all samples irrespective of their designed operating range included many configurations outside of the typical range of use of a given FREE.

We were not able to identify a clear pattern for when buckling occurred. While there appears to be some tendency for buckling under axial compression and negative end-to-end rotations, this did not hold for all samples: buckling was also observed in axial stretch and positive end-to-end rotation.
Furthermore, some samples buckled primarily at high input pressures, while others buckled at low pressures. This lack of a pattern is likely attributed to the presence of multiple different modes of buckling that depend on a sample's initial geometry, loading, kinematic state and fiber angle \citep{liu2014artery,han2013artery}, as well as defects and imperfections in the samples \citep{lee2016geometric}. A deeper investigation and classification of the observed buckling modes was not the focus of this project.

\begin{figure}
    \centering
    
    \begin{tabular}{c c}
        \begin{tabular}{c c c c}
            \begin{subfigure}[b]{1.5in}
                \centering
                \includegraphics[width=\linewidth]{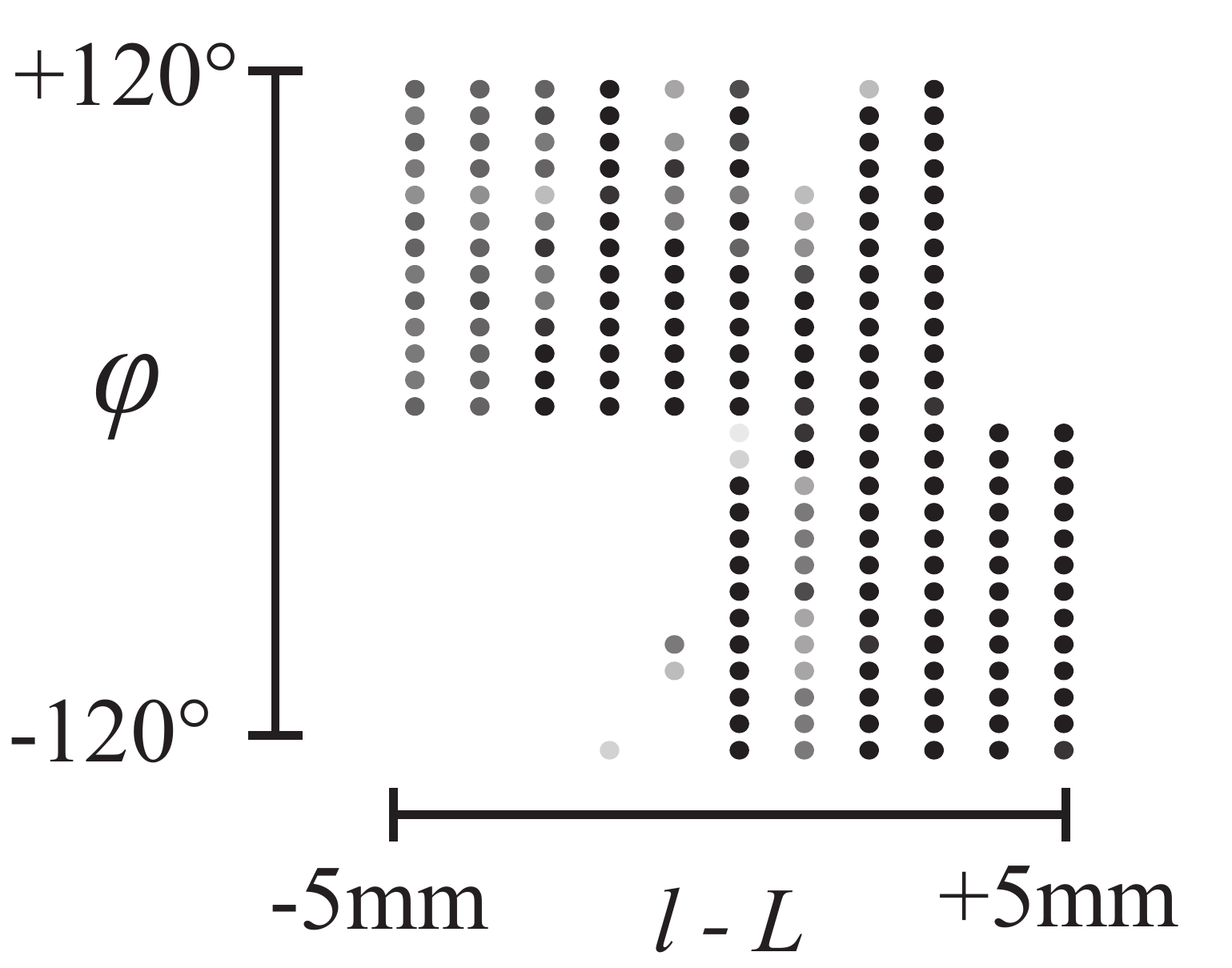}
                \caption{Sample 1 ($\Gamma = 14 \degree$)}
                \label{fig:buckled1}
            \end{subfigure} &      
        
       \begin{subfigure}[b]{1.2in}
            \centering
            \includegraphics[width=\linewidth]{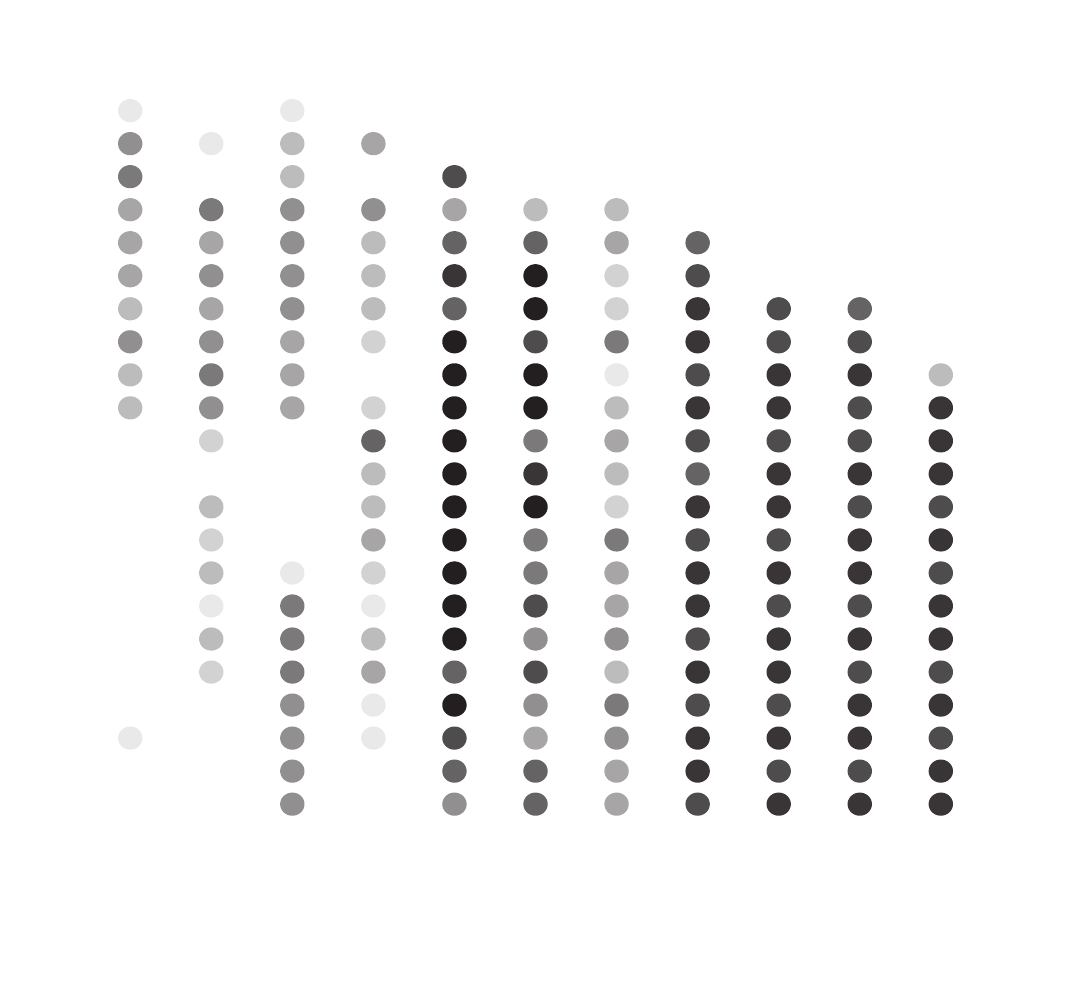}
            \caption{Sample 2 ($\Gamma = 25 \degree$)}
            \label{fig:buckled2}
        \end{subfigure} &
        
        \begin{subfigure}[b]{1.2in}
            \centering
            \includegraphics[width=\linewidth]{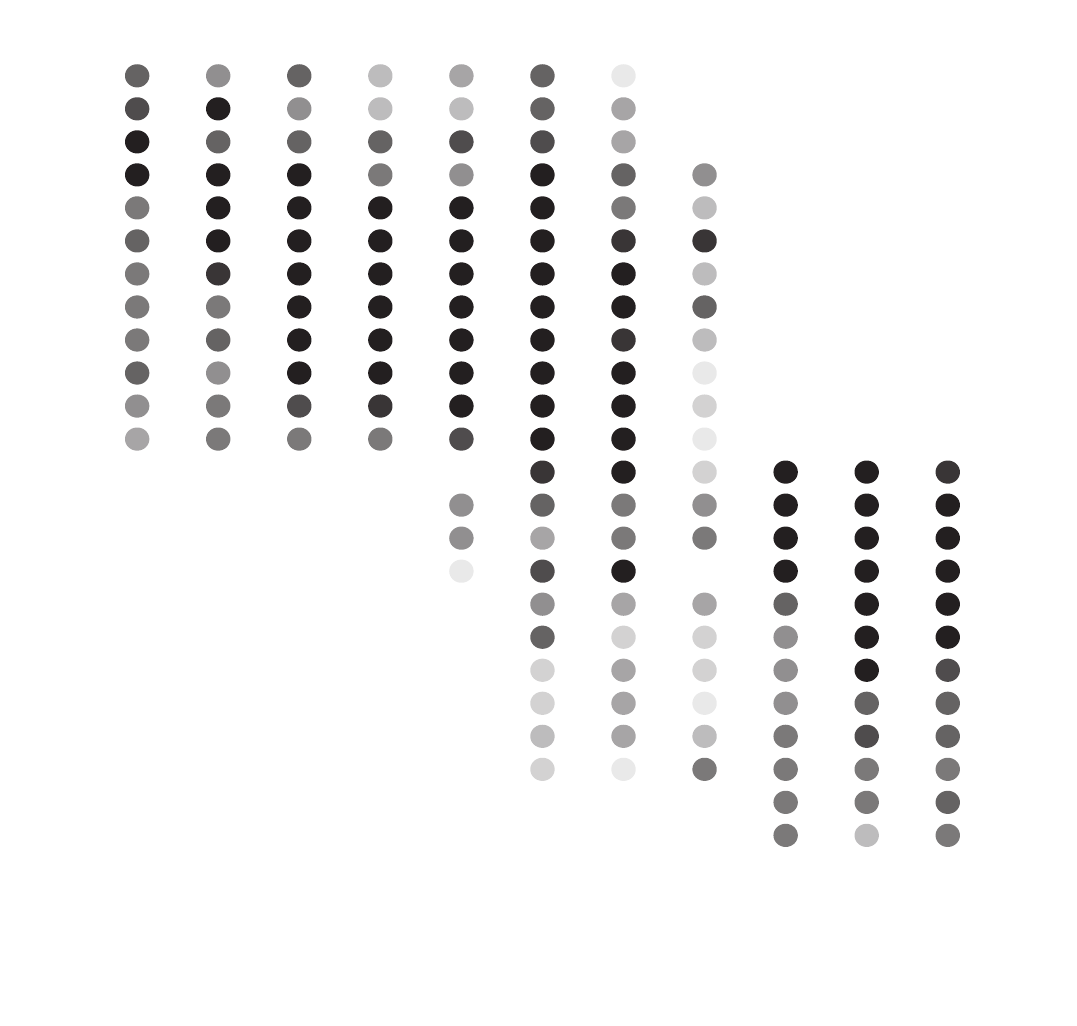}
            \caption{Sample 3 ($\Gamma = 36 \degree$)}
            \label{fig:buckled3}
        \end{subfigure} &     
        
        \begin{subfigure}[b]{1.2in}
            \centering
            \includegraphics[width=\linewidth]{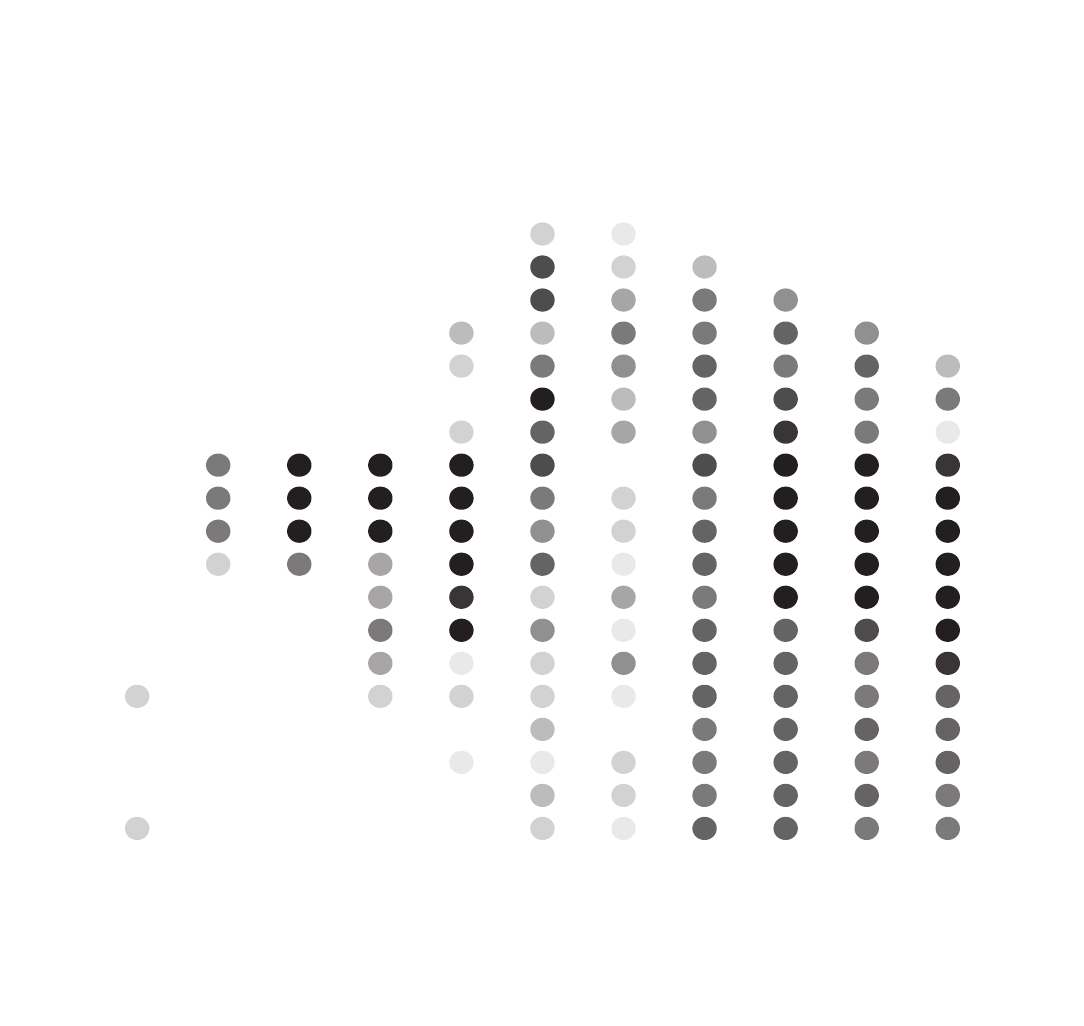}
            \caption{Sample 4 ($\Gamma = 40 \degree$)}
            \label{fig:buckled4}
        \end{subfigure}
        
        \\
        
        \begin{subfigure}[b]{1.5in}
            \centering
            \includegraphics[width=\linewidth]{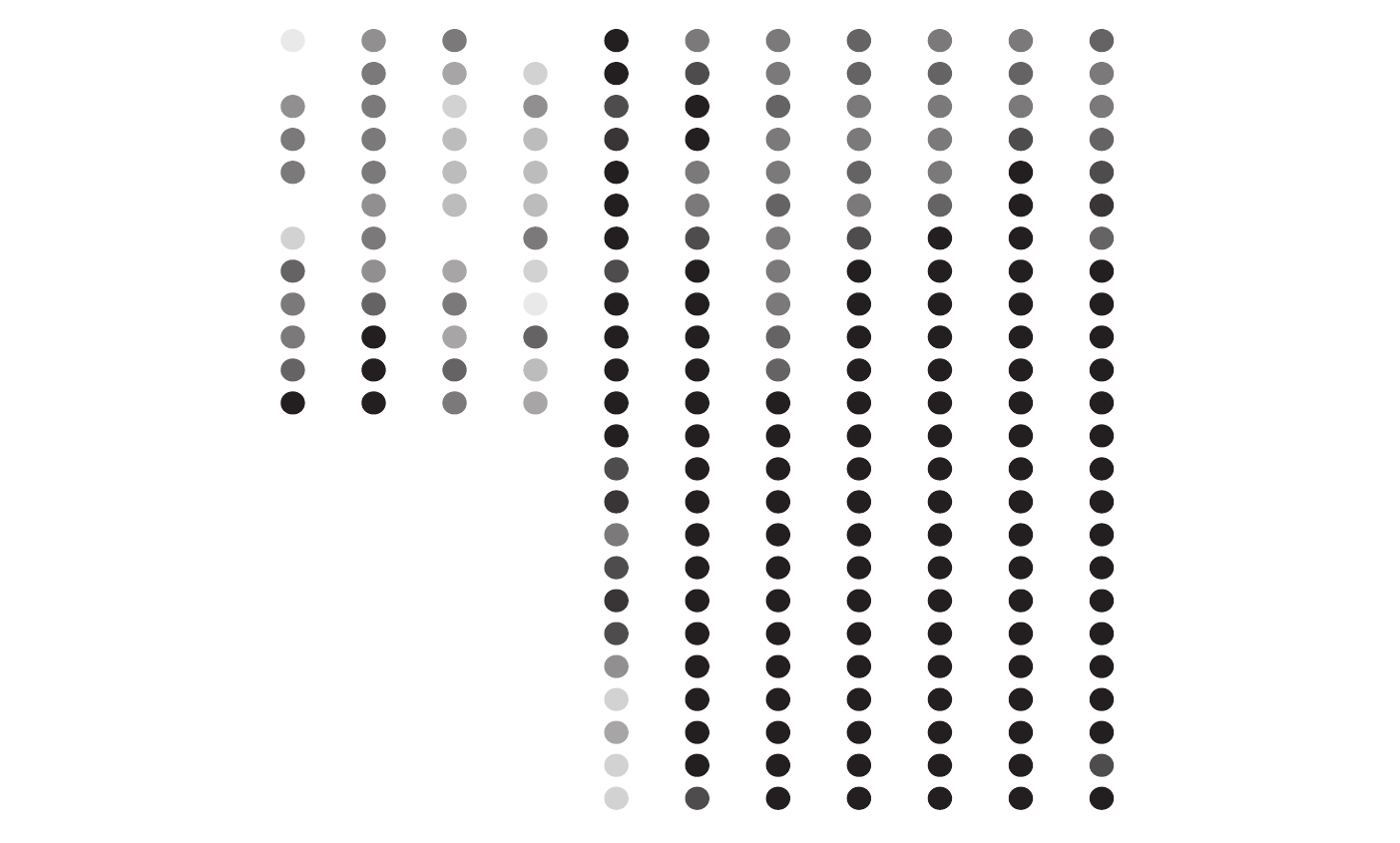}
            \caption{Sample 5 ($\Gamma = 50 \degree$)}
            \label{fig:buckled5}
        \end{subfigure} &
        
        \begin{subfigure}[b]{1.15in}
            \centering
            \includegraphics[width=\linewidth]{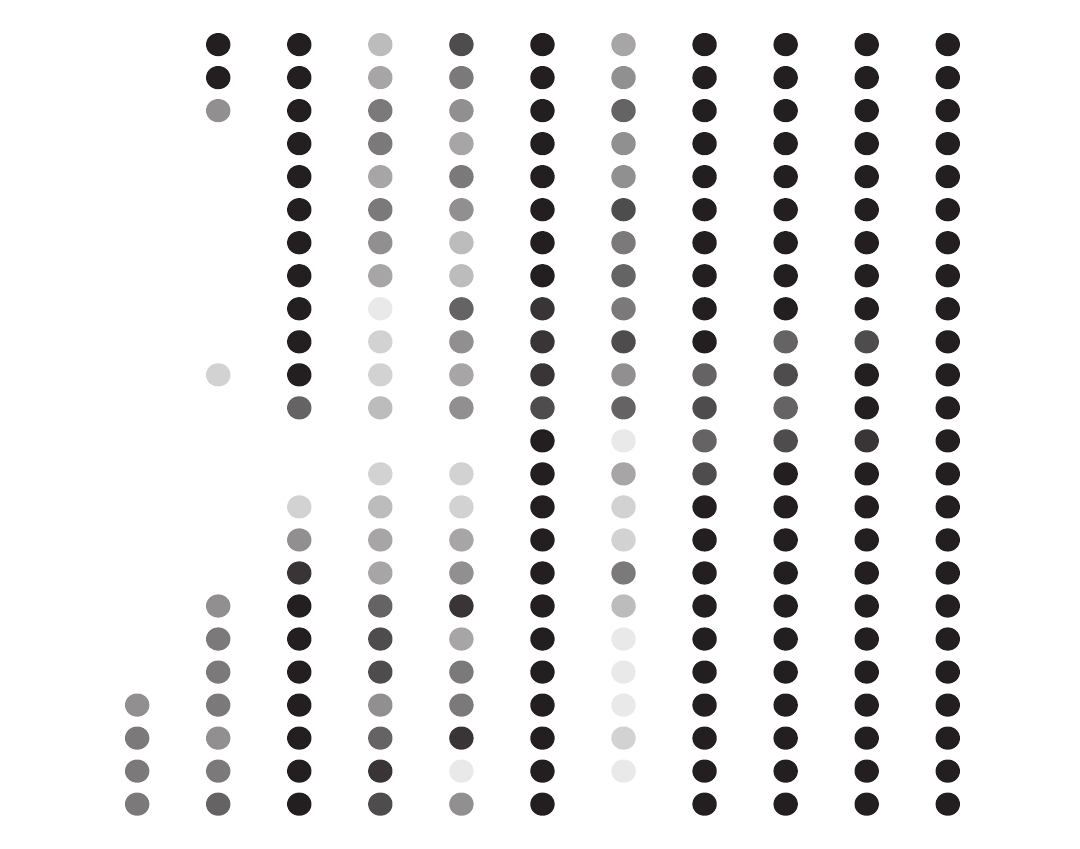}
            \caption{Sample 6 ($\Gamma = 62 \degree$)}
            \label{fig:buckled6}
        \end{subfigure} &
        
        \begin{subfigure}[b]{1.17in}
            \centering
            \includegraphics[width=\linewidth]{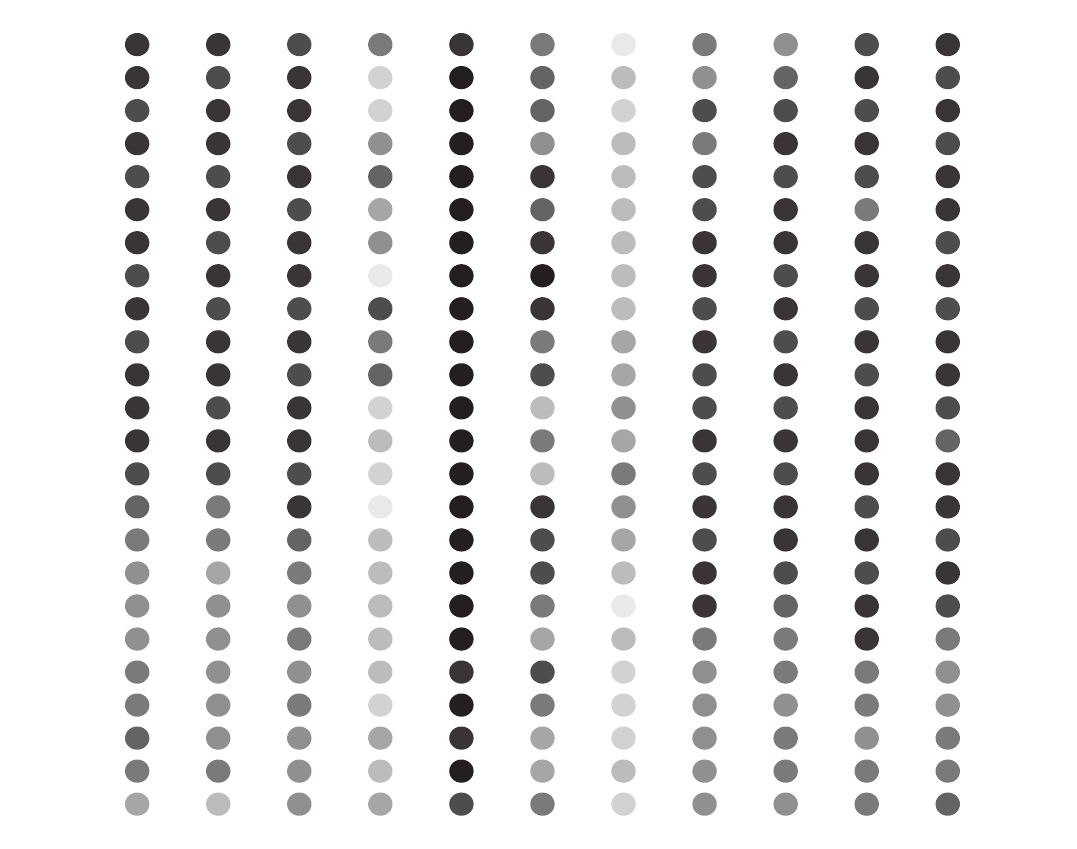}
            \caption{Sample 7 ($\Gamma = 73 \degree$)}
            \label{fig:buckled7}
        \end{subfigure} &
        
        \begin{subfigure}[b]{1.17in}
            \centering
            \includegraphics[width=\linewidth]{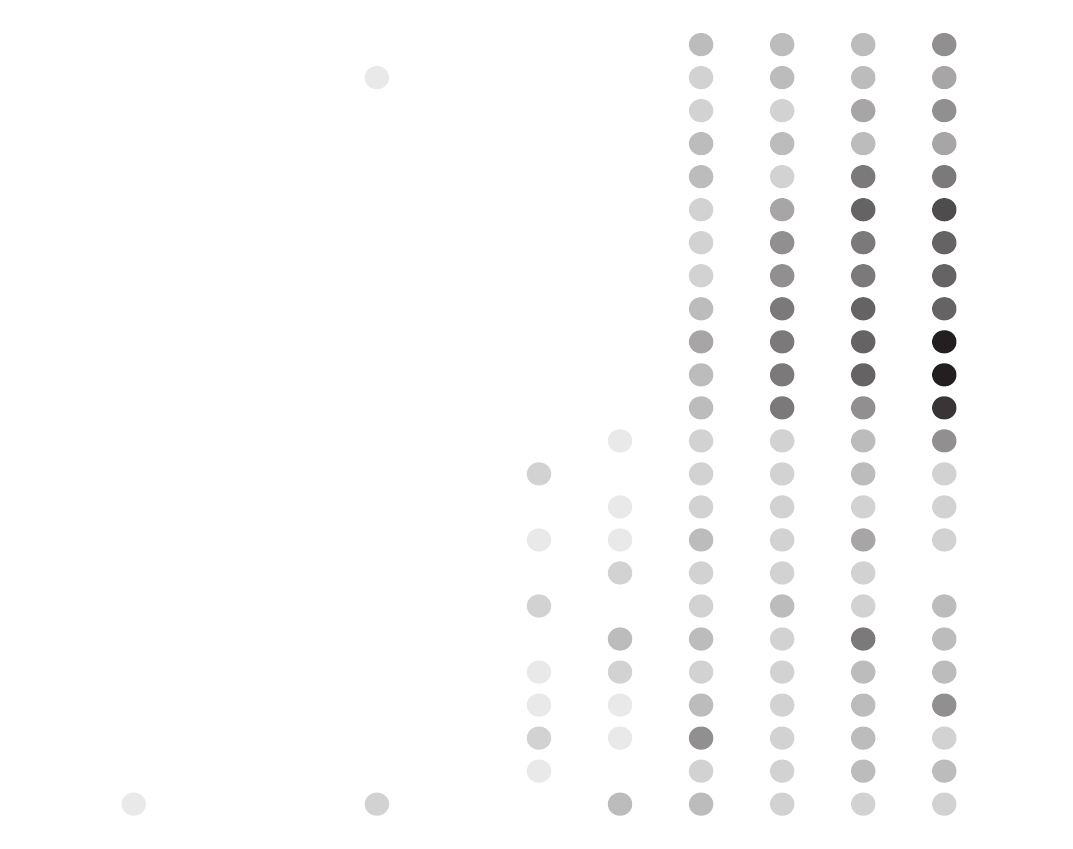}
            \caption{Sample 8 ($\Gamma = 76 \degree$)}
            \label{fig:buckled8}
            \label{fig:bucklingMaps}
        \end{subfigure}
    \end{tabular} 
    
    &
    \parbox[c]{1em}{\includegraphics[width=0.67in]{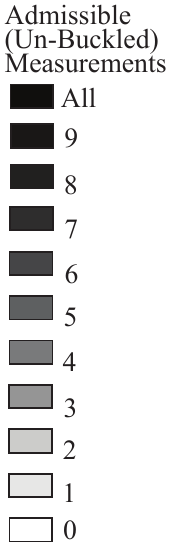}}
     
    \end{tabular}
    
\caption{
Diagrams showing admissible data (from un-buckled trials) to be used for analysis. Data are organized by kinematic state $\vec{q}$ for all 8 samples. The horizontal axis is the length change $l-L$ in mm, while the vertical axis gives the twist $\varphi$ in degrees. At each of these kinematic states, the FREE did not necessarily stay buckled or un-buckled across the pressure range imposed. The relative proportion of admissible (i.e. un-buckled) measurements at each kinematic state $\vec{q}$ is reflected by the color of the circle.}

\end{figure}

\subsection{Actuator behavior \& Model Features}
\label{sec:rawResults}

\begin{figure}
    \centering
    \includegraphics[width=3in]{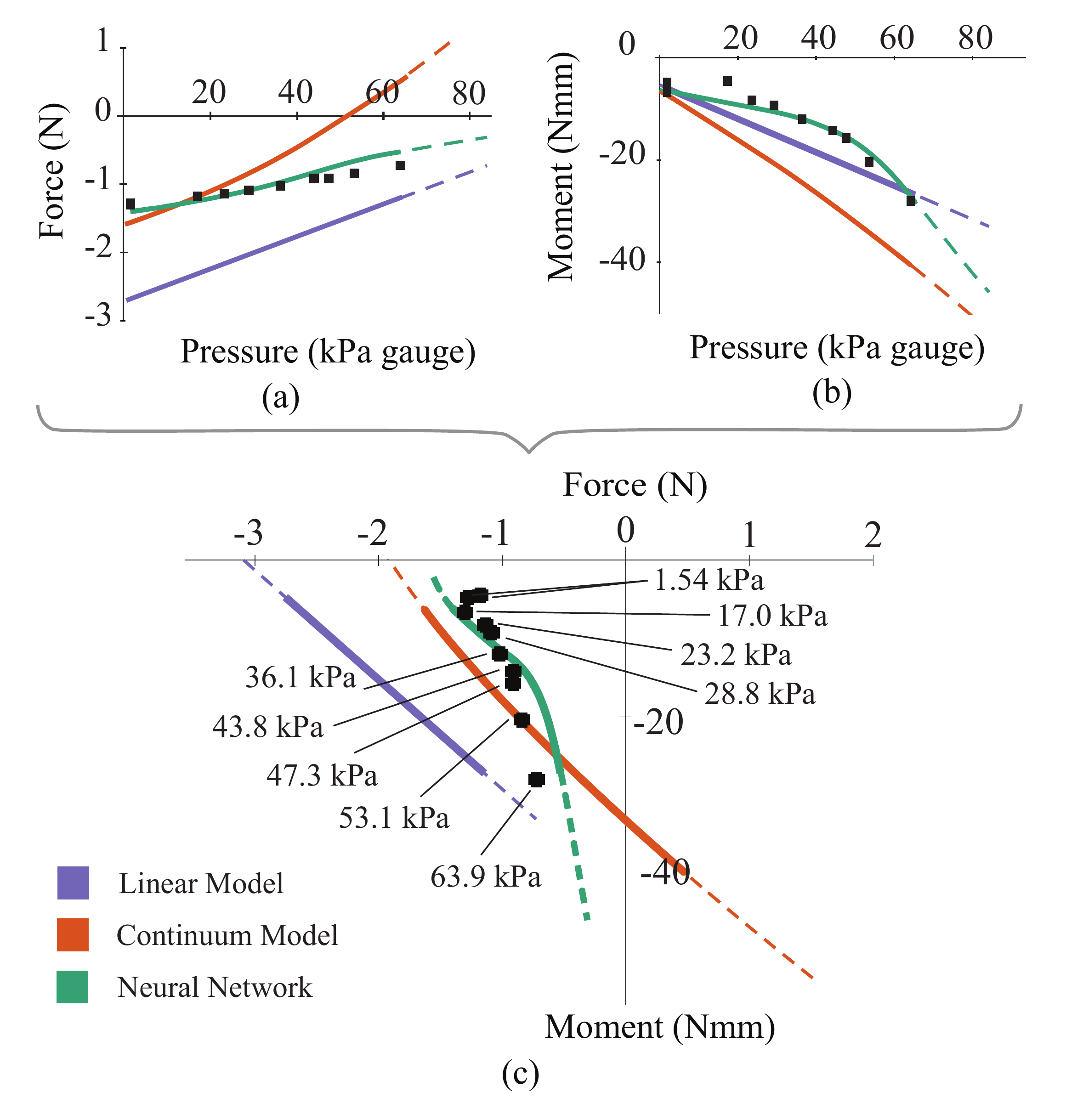}
    \caption{Force and torque models (curves) compared to measurement (black squares) for Sample 6 at $\vec{q} = [4mm \; 80\degree]$. From top left: (a) force $F$ by pressure, (b) moment $M$ by pressure, and (c) curves of $\vec{\tau} = [F M]^T$ parameterized by pressure $P$. In subfigures (a)-(c), solid lines show the predictions of $\vec{\tau}$ corresponding to the pressures $P \in \{1.54, \; 63.9\}$ in which loading measurements were taken. Dashed lines show a larger modeled range for this sample from $P = 0$ kPa (upper left end) to $P = 70.0$ kPa (lower right end).}
    \label{fig:sampleModels6}
\end{figure}

\begin{figure}
    \centering
    \includegraphics[width=4.5in]{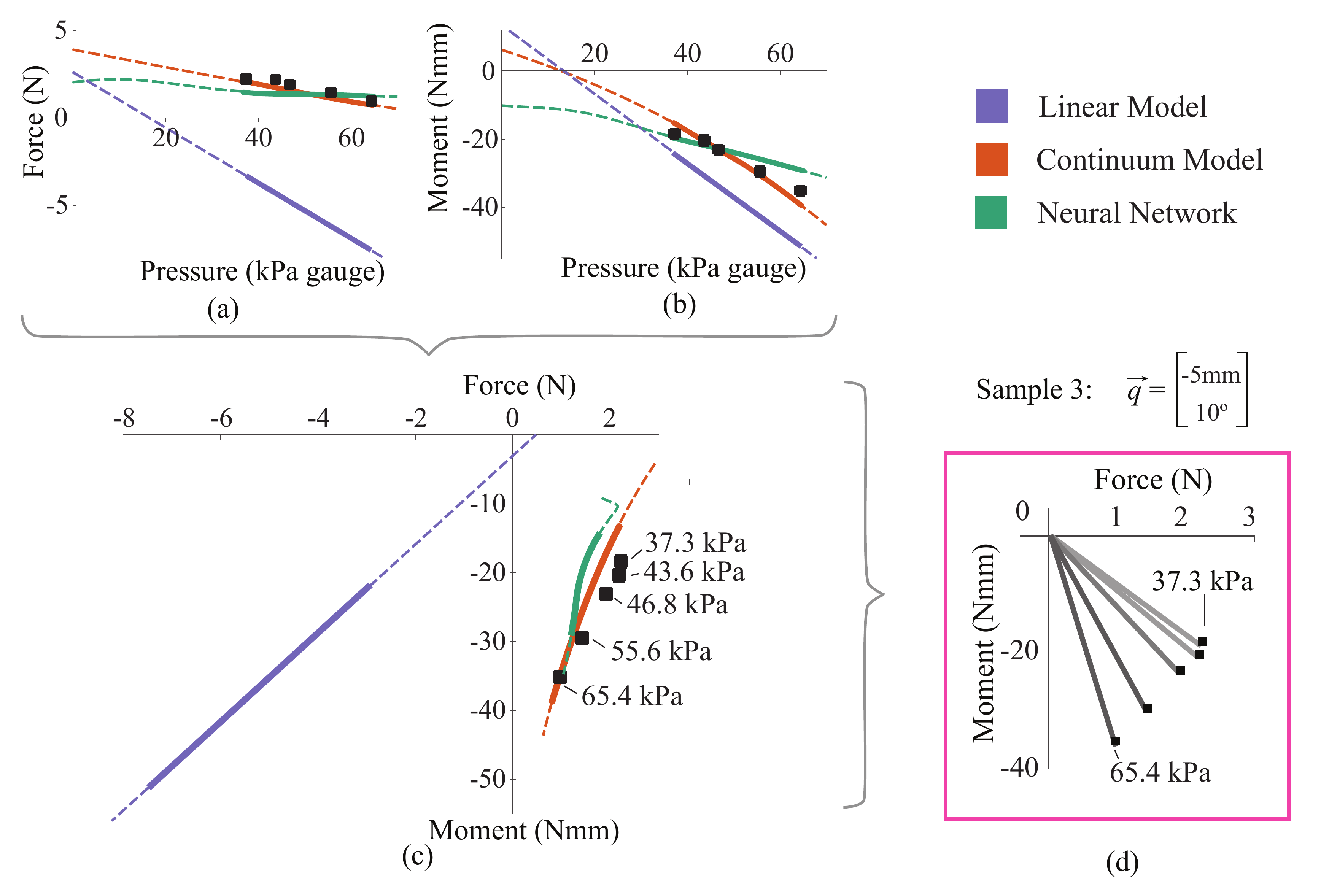}
    \caption{Force and torque models (curves) compared to measurement (black squares) for Sample 3 at $\vec{q} = [-5mm \; 10\degree]$. From top left: (a) force $F$ by pressure, (b) moment $M$ by pressure, and (c) curves of $\vec{\tau} = [F M]^T$ parameterized by pressure $P$. In subfigures (a)-(c), solid lines show the predictions of $\vec{\tau}$ corresponding to the pressures $P \in \{37.3, \; 64.4\}$ in which loading measurements were taken. Dashed lines show a larger modeled range for this sample from $P = 0$ kPa (upper right end) to $P = 69.4$ kPa (lower left ends).}
    \label{fig:sampleModels}
\end{figure}

\begin{figure}
    \centering
    \includegraphics[width=5.8in]{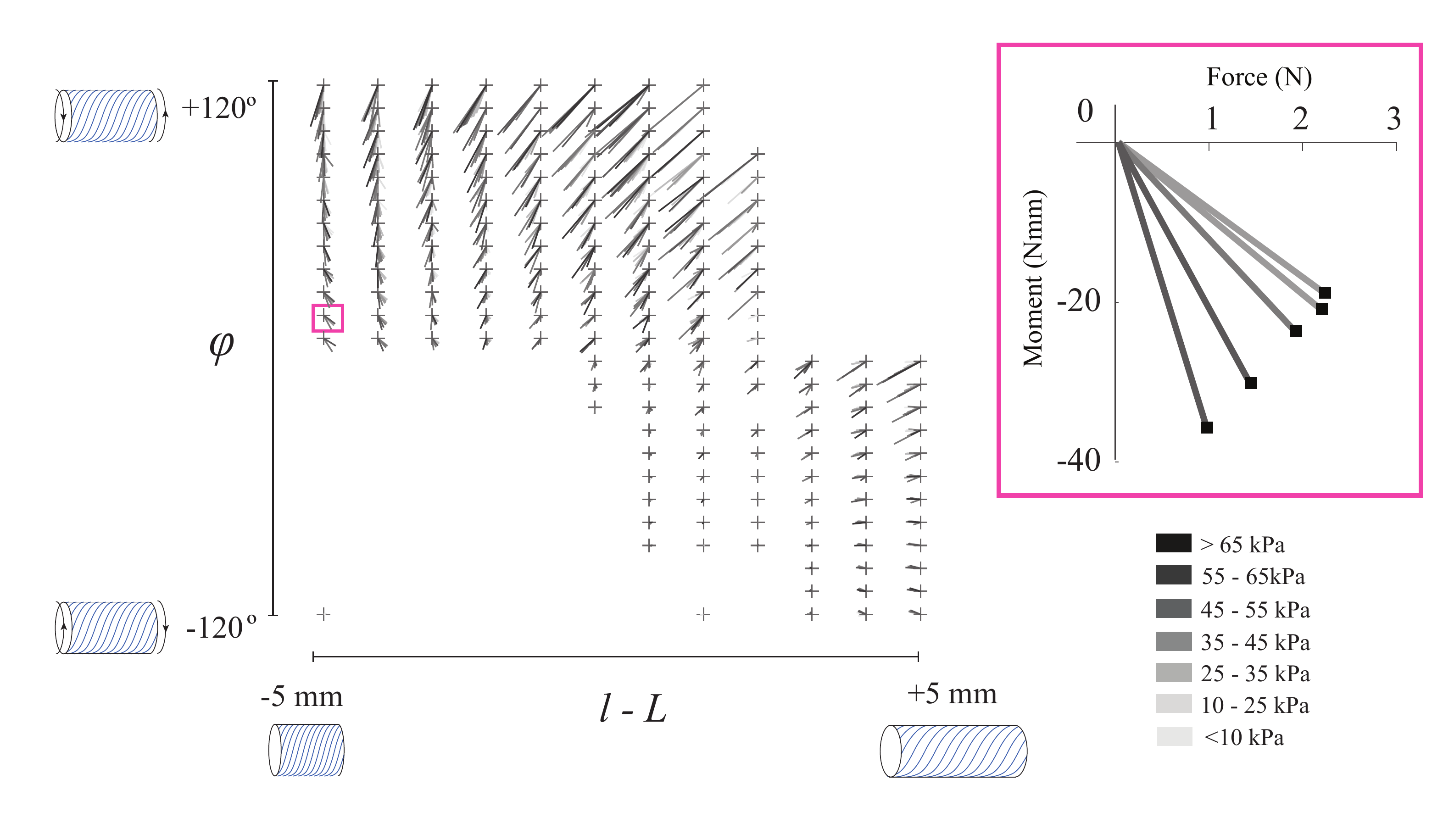}
    \caption{Raw loading data of Sample 3, shown as vectors $\vec{\tau}$ for all pressures across the space of deformation states $\vec{q}$. The pink square shows one example of how $\vec{\tau}$ changes with internal pressure at $\vec{q} = [-5mm, \; 10\degree]^T$.}
    \label{fig:rawData}
\end{figure}

To highlight the general behavior of the models, we first present a selection of data: two data sets recorded at specified kinematic states with corresponding comparisons of the modeled axial force $F$ and moment $M$ as functions of input pressure $P$ to the measurements (Figures \ref{fig:sampleModels6} and \ref{fig:sampleModels}). We then present all the recorded behavior of a particular FREE sample (Figure \ref{fig:rawData}). %show two sets of data comparing the experimental measurements of loads $\vec{\tau}$ at a fixed kinematic state $\vec{q}$ on two samples. 
 Figure \ref{fig:sampleModels6} shows the data for Sample 6 at $\vec{q} = [4mm \; 80\degree]^T$ (i.e. under axial extension and positive twist) at 10 pressure values between 1.54 kPa and 63.9 kPa. The models shown here were all trained on the data from Sample 6. Figure \ref{fig:sampleModels6}a shows the axial force $F$ arranged as a function of internal pressure $P$, and Figure \ref{fig:sampleModels6}b shows the moment $M$ as a function of $P$. Figure \ref{fig:sampleModels6}c shows the same measurements arranged to show the $F$-$M$ relationship parameterized by $P$.
Similarly, Figure \ref{fig:sampleModels} shows the data for Sample 3 at $\vec{q} = [-5mm \; 10\degree]^T$ (i.e. axial compression and positive twist) and five pressure values between 37.3kPa and 64.4kPa, with all model parameters trained on Sample 3.
In the data set of Figure \ref{fig:sampleModels}, five configurations, all at pressures lower than 37.3kPa, have been excluded from the analysis because the sample was buckled. 

The same format as Figures \ref{fig:sampleModels6}c and \ref{fig:sampleModels}c is used for the full set of measurements of Sample 3, shown in Figure \ref{fig:rawData}. Here, the $F$-$M$ relationship is shown as a vector (i.e. the vector generalized forces $\vec{\tau}$). The correspondence of the data sets is shown by the pink insert, also present in Figure \ref{fig:sampleModels}d. From the full behavior of the sample in Figure \ref{fig:rawData}, we can observe general system trends including the relative influence of kinematic state and internal pressure on the magnitude and direction of $F$ and $M$. The forces and torques produced at un-buckled configurations are characterized by behaviors due to the wall's elasticity, and behaviors due to the pressure input. Force and moment offsets %(Figures \ref{fig:sampleModels6}ab, \ref{fig:sampleModels}ab)
reflect elastic behavior of a FREE at an imposed kinematic state $\vec{q}$, at $P \approx P_{atm}$. Loading trends reflect how the loads $\vec{\tau}$ change as a function of pressure input $P$. Both of these depend on the initial design parameters $\bar{p}$. 

We did not observe significant hysteresis in this experiment. For example, in the kinematic configuration called out in Figures \ref{fig:sampleModels6} and \ref{fig:sampleModels}, loading measurements $\vec{\tau}$ at similar pressures overlap.  Further, the paths taken by the vectors $\vec{\tau}$ in Figures \ref{fig:sampleModels6}, \ref{fig:sampleModels} and \ref{fig:rawData} 
do not appear to change significantly during the pressure ascent or descent.
In general, the forces $\vec{\tau}$ produced by an actuator are highly dependent on its kinematic state $\vec{q}$. Throughout the experimental data set, larger magnitudes of $\vec{\tau}$ tend to occur at higher pressures $P$ and larger deformations $\vec{q}$. This is reflected for Sample 3, where the magnitudes of $\vec{\tau}$ in the upper right and left quadrants of Sample 3's kinematic space (Figure \ref{fig:rawData}) are larger than those in the center. Further, the largest magnitudes of $\vec{\tau}$ of Sample 3 occur in the upper right quadrant, where the kinematic states impose tension on the fibers. In the lower left quadrant, low forces occur near kinematic states where the FREE was buckled for all the pressures $P$ tested (Figure \ref{fig:buckled3}). These low forces may indicate the onset of the buckling instability, where the fibers are no longer in tension but the wall has not yet buckled. At each kinematic configuration, the direction of $\vec{\tau}_{meas}$ can vary with pressure. Comprehensive measurements for all samples at all un-buckled configurations are found in the data packet available as a supplement to this paper.

Model comparisons to these data are also shown in Figures \ref{fig:sampleModels6} and \ref{fig:sampleModels}. The shapes of the pressure-force, pressure-moment, and force-moment relations for the linear model (shown in blue) are all straight lines. The $x-$ and $y-$ intercepts of each of these straight lines are set by the experimentally determined parameters of the FREE wall stiffness matrix \textbf{K}, while the fluid Jacobian $J_V$ determine their slopes as a function of pressure $P$. The slight curves in the continuum model lines (shown in orange) reflect its geometric and strain-energy nonlinearities. The curve produced by the neural network (shown in green) does not extrapolate smoothly, but shows a hooked shape outside of the region of available data in Figure \ref{fig:sampleModels}.
In this instance, the neuron weights and activation function biases fit to the region of available data did not necessarily extrapolate in a comparable way to the physical models.

Model performance varied widely across samples, kinematic states, and input pressures. Though the curves of Figures \ref{fig:sampleModels6} and \ref{fig:sampleModels} demonstrate how each model's mathematical structure affects its behavior, they should not be used to draw conclusions about which model is most accurate across the data set. A gross performance comparison of the models organized by experimental parameter training set is given in Section \ref{sec:resModelErr}. 

\subsection{Model Comparison}
\label{sec:resModelErr}
\begin{figure}
    \centering
    \includegraphics[width=\linewidth]{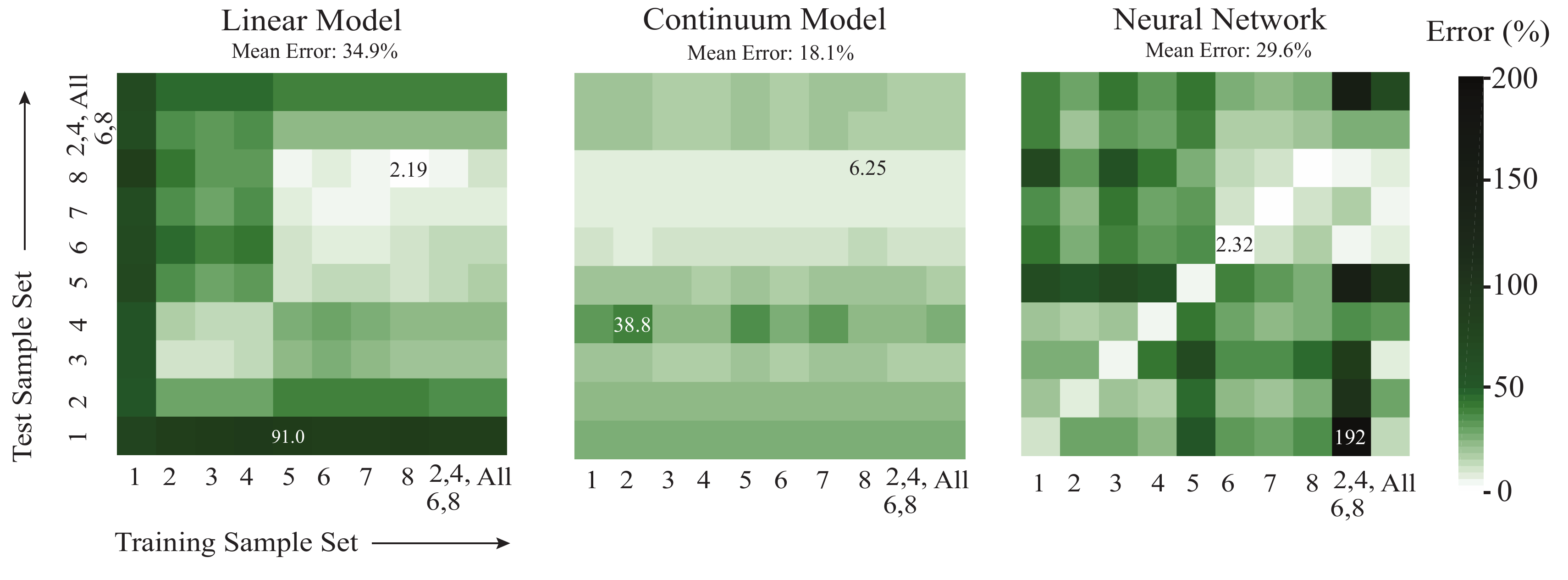}
    \caption{Normalized error, as described in Eq. 
\ref{eq:errorObj}, shown as heat maps, for all possible training-test pairs. Within each heat map, each square's color designates the error value according to the scale shown in the figure, with darker squares indicating higher error. The horizontal axis indicates which training data were used to fit parameters, while the vertical axis indicates the test set used to evaluate performance. Training and test sets were created individually for Samples 1 through 8, and for two composite sets of Samples 2, 4, 6 and 8 and all the samples, respectively. Maximum and minimum normalized errors for each model are shown in percent in the corresponding cells of the heat maps. Mean normalized error of each model is noted above its heat map.}
    \label{fig:errComposite}
\end{figure}

\begin{figure}
    \centering
    \includegraphics[width=1.3in]{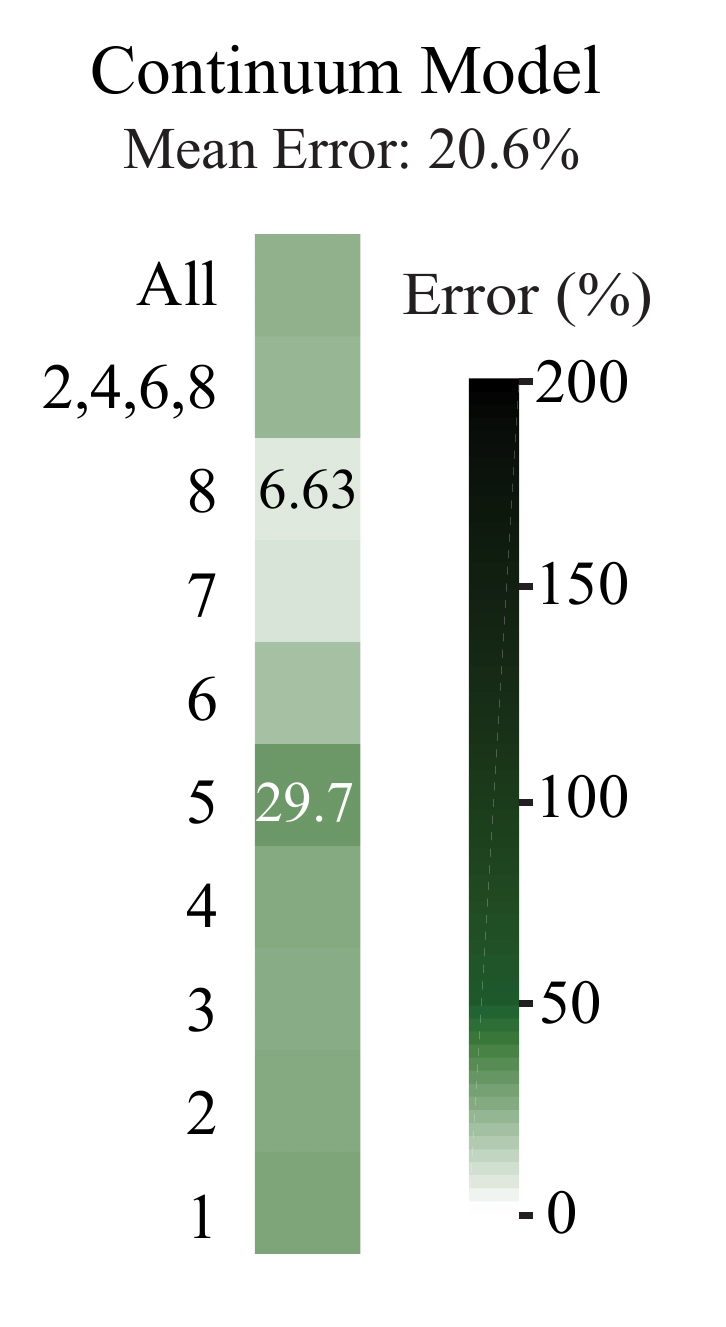}
    \caption{Normalized error of the continuum model using material properties $C_{1,2}$ determined on the individual constituent materials in \citet{sedal2018continuum}. Maximum and minimum normalized error are shown on the test set for which they occurred, and mean error is noted above the figure.}
    \label{fig:contExptFit}
\end{figure}

After partitioning the data into training and test sets, we evaluated model performance for every possible training-test pair. The result is 100 error calculations per model, which are depicted on the heat maps of Figure \ref{fig:errComposite}. Since the parameters of the continuum model may also be fit through separate experiments on the FREE's individual component materials, 10 additional error figures are shown in Figure \ref{fig:contExptFit}. 
 
The heat map of the linear model (Figure \ref{fig:errComposite}, left) is characterized by two low-error rectangular zones: a larger one at the top right and a smaller one at the bottom left. The low-error rectangular zones show regions where the linear model generalizes. When the stiffness parameters in \textbf{K} are fit to any of Samples 2 through 4, the model extrapolates relatively well in the corresponding test sets, shown by the relatively paler regions in the subplot. Similarly, models trained on Samples 5 through 8 extrapolate well across those test sets. Indeed, the lowest error achieved by the linear model is $2.19\%$ for the training and test sets of Sample 8. This is also the lowest error achieved by any of the models for any training-test pair. The separation of this heat map into regions might be due to the Sample 5 reaching the ``magic" fiber orientation \citep{goriely2013rotation} $\gamma = 54.7\degree$ that maximizes its internal volume under the linear model's assumptions. Both theory and observation show that behavior on either side of this ``magic" angle will differ as the internal pressure pushing against the sample walls leads to a volume increase. Along the bottom, top, and left edges of the heat map, bands of higher error appear; these occur near the singularity described in Section \ref{sec:lpm}. The value of twist that gives a singular configuration of Sample 1  occurs at $\varphi = -\frac{L}{R_o}\tan(\Gamma) = 122.4\degree$; some kinematic states of Sample 1 in this experiment nearly overlap with the singularity. Higher error then occurs in cases where the training set or the test set of from Sample 1. The ``All" test set, which includes Sample 1, also has higher error. Here, the fitted parameters of the stiffness matrix \textbf{K} differed by a maximum factor of 686 across all samples ($k_a$ trained from Sample 8 vs. $k_a$ trained from Sample 1), but by a maximum factor of 4.7 within the range of Samples 5 through 8 ($k_a$ of Sample 8 vs. $k_a$ of Sample 5) , and a maximum factor of 3.2 between Samples 2 and 4 ($k_a$ of Sample 4 vs. $k_a$ of Sample 2).

The heat map of the continuum model (Figure \ref{fig:errComposite}, center) is characterized by its low variation in error: no regions of the map are particularly bright or dark. Instead, regions of relatively high and low error of the continuum model form horizontal ``bands" corresponding with the test set used. For example, the error of Samples 7 and 8 is lower than the minimum error of Sample 2, no matter to which measurements the parameters $C_{1,2}$ were fit. A particularly dark band occurs when the test set used is Sample 4, but the maximum error of this model ($38.8\%$ for the model parameters fit from the Sample 2 training set and tested on the Sample 4 test set) is still substantially lower than the maximum error of the other models. This lower range of error is further reflected in a the lower range of parameters fit across the design space. Here, the parameters fit within the continuum model differed across all samples by a maximum factor of 3.6, ($C_1 \in \{0.94 \times 10^5, \;3.39 \times 10^5\}$ for Sample 8 and $C_2\in\{0.996 \times 10^6, \;1.10\times 10^6 \}$ for Sample 5).

The continuum model was also evaluated with the physical parameters $C_{1,2}$ of individual constituent materials (i.e. elastomer and fiber) in \citet{sedal2018continuum}. These are shown in Figure \ref{fig:contExptFit}. The minimum error here ($6.63\%$) is similar to the minimum error achieved in the earlier heat map ($6.25\%$), while the maximum error is slightly lower ($29.7\%$ vs $38.8\%$). Material constants used here ($C_1 = 5 \times 10^5$ Pa and $C_2 = 1 \times 10^6$ Pa) and the material constants identified in our experiment were within an order of magnitude.

The heat map of the neural network (Figure \ref{fig:errComposite}, right) is characterized by its bright, low-error diagonal. Performance of the neural network is especially strong when training and test data from the same sample are used, shown by the lower errors on the diagonal of the heat map. Indeed, the neural network produces the lowest error of any model for the test sets of Samples 1 through 7, though the linear model has the lowest error for Sample 8. Further, the relatively lower errors of Samples 6, 7, and 8 persist in the neural network model, shown by the brighter region of the heat map in the upper right hand quadrant. Elsewhere, the neural network has much higher off-diagonal errors.

 The neural network heat map in Figure \ref{fig:errComposite} provides evidence of over-fitting when aggregate training sets are used. In the training set of a single sample, test conditions (kinematic states $\vec{q}$ and the input pressure $P$) are varied but the design parameters $\bar{p}$ are the same. As a result, weights and biases fit to $\bar{p}$ are of value zero for all neural networks trained on a single sample. In aggregate training sets, however, the design parameters $\bar{p}$ vary according to Table \ref{table:fiber_angles}, forming a relatively sparse space of design parameters compared to the amount of testing conditions. Here, then, the weights and biases fit to $\bar{p}$ are likely to be important. Since these weights and biases have no physical meaning, they may not extrapolate or interpolate appropriately to initial fiber orientations, lengths, and radii outside of the trained values. When trained on Samples 2, 4, 6, and 8, the neural network performed relatively poorly at predicting the behavior of Samples 1, 3 and 5, with $192\%$, $89.8\%$, and $145\%$ error respectively. The influence of this effective addition of parameters is also seen by comparing neural network error on Sample 5: trained on Samples 2,4,6, and 8, the neural network gives $145\%$ load prediction error on Sample 5. This is higher than its error on Sample 5 when trained only on Sample 2 ($54.0\%$), only on Sample 4 ($56.9\%$), only on Sample 6 ($39.6\%$), or only on Sample 8 ($25.9\%$). Indeed, error here was lower when we effectively ignored the differences in design parameters than when we tried to fit to them. 

%=============================================================================
\section{Discussion}
\label{sec:discussion}
Though specific modeling techniques have been validated against realized soft robotic systems in the past, a model comparison across classes of models on a common system has not been made. To address this gap, we developed and compared distinct models that relate the loading and deformation of Fiber-Reinforced Elastomeric Enclosures (FREEs). Three static models were developed and evaluated: a linear lumped-parameter model, a nonlinear continuum mechanical model, and a neural network. We compared predictions of the three models to 12,575 loading measurements that form the broadest (to the authors’ knowledge) data set of FREE loading. The data set is spanning eight varied designs under over hundreds of distinct kinematic deformations and input pressures. Together, these evaluations of model performance enable a comparison of the models for peak performance, generalizability, and system identification effort.

We begin by evaluating the peak performance of each model. Intuition may suggest that each model would perform best when its system identification and evaluation use measurements from the same sample --- i.e., the diagonal of the heat maps of error in Figure \ref{fig:errComposite}. The neural network heat map displays this pattern in a pronounced way, with a brighter diagonal than the linear or continuum models. The average neural network error along the diagonal for the eight samples is $5.51\%$, while the same average error is $21.4\%$ for the linear model and $16.87\%$ for the continuum model. In most cases along the diagonal, the neural network achieves the lowest error ($11.5\%, 7.45\%, 6.00\%, 5.96\%, 5.97\% 2.32\%$ and $2.37\%$ for Samples 1-7 respectively). The linear model achieves the lowest error ($2.19\%$) on Sample 8. Bounded high performance could be especially useful for building models that support control schemes for already-realized soft systems in clearly defined environments. \citet{giorelli2015neural} use the neural network for this task, learning inverse statics of their soft tentacle system from measurements taken across all of its possible kinematic configurations.

Having shown that peak performance of a model may be confined, we now consider how well the models generalize across the eight samples. A model that generalizes well maintains its performance even as its training and test set vary, resulting in a uniform color across its heat map. The continuum model follows this pattern, having the smallest error range of all models across the training-test pairs ($32.6\%$) , and the lowest average error of $18.1\%$. In contrast, the linear model has an error range of $88.8\%$ and a mean error of $34.9\%$, and the neural network has an error range of $189.7\%$ and a mean error of $29.6\%$. Instead of contrasting diagonals and off-diagonals, the continuum model heat map has horizontal ``bands" of similar color, with Samples 6-8 performing best. These results suggest that the continuum model’s nonlinear mathematical structure is crucial to its performance: neither its parameters $C_{1,2}$ (as noted in Section \ref{sec:resModelErr}) nor its range of error differ as much as those of the other models across training sets. 

The linear model also generalizes, but does so in rectangular regions, described in greater depth in Section \ref{sec:resModelErr}. This type of linear model has successfully predicted loads for specified parallel combinations of FREEs \citep{bruder2018force}, but faces limitations due its singularity and the delineation of the training-test pairs for which it performs best. 

When training and test samples differ, the neural network fails to generalize. This behavior is indicated by the bright diagonal and contrasting darker off-diagonals on the neural network heat map. The neural network's higher amount of experimentally determined parameters (62) and their lack of physical meaning are the likely cause of this behavior. Since the network is agnostic to any physical assumptions and the parameters carry no particular physical meaning, it may enhance its performance by learning systematic error associated with each individual sample (e.g. fiber irregularity), or capturing phenomena neglected in the simple first principles-based models. These phenomena, however, may not occur again or in the same ways on a different sample.

In general, the performance of all models presented here is best for Samples 6-8 for any training set used. The key distinction of these samples from Samples 1-5 is in the winding angle $\Gamma$ of the helical fiber (Table \ref{table:fiber_angles}), though the initial dimensions also differ slightly. Improved performance may, then, may be due to the relatively more important role of a high-angle fiber in constraining the radial expansion of the FREE, making it act more like a cylindrical piston and better constraining the radial expansion of the FREE. Or, higher fiber angles may be easier to manufacture consistently. This phenomenon should be the subject of future study.

Though our results show that the continuum model's structure is crucial in its performance, it is not clear which aspect of its nonlinear structure is most important. In contrast to the nonlinear strain energy models presented here, \citet{coevoet2017software} show that finite element material models with nonlinear actuator deformations but linear stress-strain relationships have sufficient accuracy to enable control of a PneuNet-like soft robot, a soft cable robot, and a compliant mechanism. To our knowledge, a continuum model structure with linear material and nonlinear deformation assumptions has not been evaluated systematically for fiber-reinforced actuators; studying how structural aspects of continuum models of FREEs affect their behavior should be the subject of future work.

Finally, we compare the system identification effort necessary in each of the models. A model with more experimental parameters is likely to require more dense experimental data, and hence a longer and more exacting data collection scheme, to perform well.  In contrast, a model with an appropriate mathematical structure may perform well with fewer experimentally determined parameters and less data collection. The linear model and the continuum model have 3 and 2 experimentally determined parameters respectively, contrasting with the neural network which has 62 experimentally determined parameters. Even when the linear and continuum model fail, they do not reach error figures as high as the maximum errors of the neural network. As indicated by Figures \ref{fig:sampleModels6} and \ref{fig:sampleModels}, the physical assumptions on which these models are built give them mathematical structure which enables extrapolation across kinematic states and input pressures. Figures \ref{fig:errComposite} and \ref{fig:contExptFit} show that they can also extrapolate across designs. The continuum model has the most flexibility in system identification. This is because the parameters $C_{1,2}$ have physical meaning that relates to the stiffnesses of a FREE's constituent materials, rather than its design. As shown by the comparable errors of the continuum model in Figures \ref{fig:errComposite} and \ref{fig:contExptFit}, a roboticist could fit $C_{1,2}$ for any existing FREE or from un-assembled constituent materials, and obtain comparable model performance as they might on the exact system that they plan to use. For the linear model, generalization occurs in bounded regimes (shown by the brighter rectangles of the linear model heat map in Figure \ref{fig:errComposite}): roboticists, then, are able to perform system identification for this model from a limited choice of other, assembled FREE designs. 

The key way to improve the performance of the linear and continuum models is to incorporate additional or refined physical phenomena. These new phenomena could be specific to an experiment or context, including interactions between our sample and test bed or the effect of defects in the FREE wall. It is possible that for some phenomena to be incorporated into a model, new experimentally determined parameters will need to be added. And, as parameters are added, roboticists will presumably have to trade between model performance and expedient system identification. However, the results in this paper suggest that an appropriate mathematical structure can reduce the need for experimental parameters.

The neural network is likely to require the strongest system identification effort. This paper has shown that the same neural network architecture, with the same number of neurons, can at once outperform first principles-based models and under-perform by over-fitting, even when trained on a larger quantity of measurements (i.e. in the composite training set of Samples 2, 4, 6, and 8). As noted in Section \ref{sec:resModelErr}, the neural network has a strong performance when trained on data for a single sample with a dense array of kinematic states and input pressures, but over-fits when trained on multiple FREE samples, which more sparsely occupy the available design space. Roboticists may need to carefully collect training data that are sufficiently dense and that span the entire set of possible inputs, kinematic states, and designs. Without the availability of dense enough data, roboticists may tune a neural network in several ways; a detailed exploration of this is an avenue for future work.

Limitations of our study are discussed below. We chose three models ---a linear lumped-parameter model, a continuum model, and a neural network--- to span a wide space of first principles-based (for the linear and continuum models) and data-driven (for the neural network) techniques. The linear and continuum models are the simplest of their respective classes, and any improved versions of these will inherit their core structure from the models presented here. Conclusions we draw about these classes of models are likely to hold for refined model versions even as performance may improve. In contrast, the neural network is not equally representative of the broad class of data-driven models. It is a popular choice in recent soft robotics modeling work, but its core features are not necessarily inherited by other data-driven techniques. Other data-driven methods relying on fundamentally different mathematical structures may be better at capturing the FREE's underlying physics. \citet{sunderhauf2018limits} note that deep learning techniques have ``been shown to learn predictive physics-based models.''
\citet{bruder2018nonlinear} show the potential of the Koopman operator in modeling parallel structures of FREEs and \citet{satheeshbabu2018open} uses deep reinforcement learning to model a manipulator made from FREEs. 

Further, we chose to study FREEs specifically because they share core structural features with both PneuNet-type soft robots and soft cable robots. Despite these similarities, the modeling styles presented here may have different behaviors when applied to other actuator styles. A similar model comparison on non-FREE actuators is an avenue for future work. 

The static nature of the experimental validation is another limitation. Though viscoelastic behavior was not evident in this experiment, it may still play a role at smaller or larger time-scales than the one presented here. Further, the addition of significant inertial forces to an experiment would clearly change the loading outcomes. A new experiment would be required for a study of the dynamic behavior of FREEs, but all of the models presented here could be adapted to include dynamics. The linear and continuum models here may be extended to include dynamics \citep{scarborough2012fluidic}, and the neural network could be fit on experimental data that includes dynamic phenomena. Having demonstrated the strengths and weakness of these models on predicting the static behavior of FREEs, a dynamic investigation should be the next step.

The work presented here provides a broad experimental benchmark enabling comparison of distinct modeling styles found across soft robotics literature. While the modeling approaches presented here have been previously proven on a variety of individual, fully functional soft robotic systems, a comparison on an expansive common data set had not yet been realized. The results shown here confirmed some behaviors already suggested by intuition – e.g. that models with a higher number of experimentally determined parameters may also have the highest peak performance – but also uncovered trends of model performance and physical system behavior across the fiber-reinforced soft actuator design space that are not observable in isolated cases. This broad understanding can help roboticists in various phases of soft system development and validation: our demonstration of the strengths, weaknesses, and failure points of each of these models can guide model choice articulated systems of FREEs or soft actuators similar to FREEs, guide design as roboticists find which schemes best suit their modeling capabilities, and inform the development of a next generation of improved soft actuator models.

%=============================================================================

\section*{Acknowledgements}
This work was funded by the Toyota Research Institute (TRI). We thank our liaison at TRI, Alejandro Castro, for helpful feedback throughout the data gathering and wiritng process. We would also like to thank Prof. Keith Buffinton (Bucknell University), Daniel Bruder, Nils Smit-Anseeuw (University of Michigan) and the members of FlexLab$@$EPFL for helpful comments, and Daniel Haley for help with experiments. %The data set can be downloaded at https://umich.box.com/s/xwhjvwidblat4dwliyc7qtosxbultnwm.

\bibliographystyle{abbrvnat}
\bibliography{References}

\end{document}